\documentclass{article}

\usepackage{graphicx}
\usepackage[latin1]{inputenc}

\usepackage{amsmath,amssymb,amsthm}

\newcommand{\G}{\mathcal{G}}
\newcommand{\V}{\mathcal{V}}
\newcommand{\E}{\mathcal{E}}
\newcommand{\F}{\mathcal{F}}

\renewcommand{\>}[1]{\mathbf{#1}}
\newcommand{\x}{\>x}

\renewcommand{\P}{\mathbb{P}}
\renewcommand{\E}{\mathbb{E}}
\newcommand{\bel}{b}

\DeclareMathOperator{\atanh}{atanh}

\newtheorem{prop}{Proposition}[section]

\newcommand\egaldef{\stackrel{\mbox{\upshape\tiny def}}{=}}
\newcommand{\1}{\leavevmode\hbox{\rm \small1\kern-0.35em\normalsize1}}
\newcommand{\ind}[1]{\1_{\{#1\}}}
\def\DD{\displaystyle} 

\begin{document}

\title{Learning Multiple Belief Propagation Fixed Points for Real Time Inference}

\author{Cyril Furtlehner
  \thanks{INRIA Saclay -- LRI, Bat. 490, Université Paris-Sud -- 91405
  Orsay cedex (France)}
\and Jean-Marc Lasgouttes
  \thanks{INRIA Paris Rocquencourt -- Domaine de Voluceau B.P.\ 105 -- 78153 Le Chesnay cedex (France)}
\and Anne Auger\footnotemark[1]}
\maketitle   

\begin{abstract}
In the context of inference with expectation constraints, we propose
an approach based on the ``loopy belief propagation'' algorithm
(\textsc{lpb}), as a surrogate to an exact Markov Random Field
(\textsc{mrf}) modelling. A prior information composed of correlations
among a large set of $N$ variables, is encoded into a graphical model;
this encoding is optimized with respect to an approximate decoding
procedure (\textsc{lbp}), which is used to infer hidden variables from
an observed subset.  We focus on the situation where the underlying
data have many different statistical components, representing a
variety of independent patterns. Considering a single parameter family
of models we show how \textsc{lpb} may be used to encode and decode
efficiently such information, without solving the NP hard inverse
problem yielding the optimal \textsc{mrf}. Contrary to usual
practice, we work in the non-convex Bethe free energy minimization
framework, and manage to associate a belief propagation fixed point to
each component of the underlying probabilistic mixture. 
The mean field limit is considered and yields an exact
connection with the Hopfield model at finite temperature and steady
state, when the number of mixture components is proportional to the number of variables. 
In addition, we provide an enhanced learning procedure, based on a
straightforward multi-parameter extension of the model in conjunction
with an effective continuous optimization procedure. This is performed
using the stochastic search heuristic \textsc{cmaes} and yields a
significant improvement with respect to the single parameter basic
model.
\end{abstract}

\section{Introduction}\label{sec:intro}

Prediction or recognition methods on systems in a random environment
have somehow to exploit regularities or correlations, possibly both
spatial and temporal, to infer a global behavior from partial
observations. For example, on a road-traffic network, one is
interested to extract, from fixed sensors and floating car data, an
estimation of the overall traffic situation and its
evolution~\cite{FuLaFo}. For image recognition or visual event
detection, it is in some sense the mutual information between
different pixels or sets of pixels that one wishes to exploit. The
natural probabilistic tool to encode mutual information is the Markov
Random Field (\textsc{mrf}), which marginal conditional probabilities
have to be computed for the prediction or recognition process.

The inference problem (with expectation constraints~\cite{HeOp}) that
we want to address is stated as follows: the system is composed of
discrete variables $\x=\{x_i,i\in \V\}\in\{1,\ldots,q\}^\V$ for which
the only known statistical information is in the form of marginal
probabilities, $\hat p_a(\x_a)$ on a set $\F$ of cliques $a\subset
\V$. Such marginals are typically the result of some empirical
procedure producing historical data. Based on this historical information, 
consider then a situation where some of the variables are observed,
say a subset $\x^* =\{x^*_i,\ i\in \V^*\}$, while 
the other one, the complementary set $\V\setminus\V^*$, remains hidden.
What prediction can be made concerning this complementary set, and how
fast can we make this prediction, if we think in terms of real time
applications, like traffic prediction for example?

Since the variables take their values over a finite set, the marginal
probabilities are fully described by a finite set of correlations and,
following the principle of maximum entropy distribution of
Jaynes~\cite{Cover}, we expect the historical data to be best encoded
in a \textsc{mrf} with a joint probability distribution of $\x$ of the
form
\begin{equation}\label{eq:joint}
\P(\x) = \prod_{i\in\V}\phi_i(x_i) \prod_{a\in\F}\psi_a(\x_a).
\end{equation}
This representation corresponds to a factor graph~\cite{Kschi}, where
by convenience we associate a function $\phi_i(x_i)$ to each variable
$i\in\V$ in addition to the subsets $a\in\F$, that we call
``factors''. $\F$ together with $\V$ define the factor graph $\G$,
which will be assumed to be connected.

There are two main issues:
\begin{itemize}
\item \emph{inverse problem}: how to set the parameters of
  (\ref{eq:joint}) in order to fulfill the constraints imposed by the
  historical data?
\item \emph{inference}: how to decode (in the sense of computing marginals) in the most efficient
  manner---typically in real time---this information, in terms of
  conditional probabilities $\P(\x|\x^*)$?
\end{itemize}
Exact procedures generally face an exponential complexity problem both
for the encoding and decoding procedures and one has to resort to
approximate procedures~\cite{WeTe}. The Bethe
approximation~\cite{Bethe}, which is used in statistical physics
consists in minimizing an approximate version of the variational free
energy associated to~(\ref{eq:joint}). In computer science, the belief
propagation \textsc{bp} algorithm~\cite{Pearl} is a message passing procedure
that allows to compute efficiently exact marginal probabilities when
the underlying graph is a tree. When the graph has cycles, it is still
possible to apply the procedure (then referred to as \textsc{lbp}, for ``loopy
belief propagation''), which converges with a rather good accuracy on
sufficiently sparse graphs. However, there may be several fixed
points, either stable or unstable. It has been shown that these points
coincide with stationary points of the Bethe free energy~\cite{YeFrWe}
which is defined as follows:
\begin{align}\label{def:bfe}
F(\bel) &= -\sum_{a\in\F}\sum_{\x_a}\bel_a(\x_a)\log\psi_a(\x_a)-\sum_{i\in\V}\sum_{x_i}
\bel_i(x_i)\log\phi_i(x_i)\nonumber\\
&+\sum_{a\in\F}\sum_{\x_a}\bel_a(\x_a)\log \bel_a(\x_a)+\sum_{i\in\V}\sum_{x_i}
(1-d_i)\bel_i(x_i)\log \bel_i(x_i).
\end{align}
In addition, stable fixed points of \textsc{lbp} are local minima of
the Bethe free energy~\cite{Heskes4}. The question of convergence of
\textsc{lbp} has been addressed in a series of
works~\cite{Tatikonda02,MooijKappen07,Ihler} establishing
conditions and bounds on the \textsc{mrf} coefficients for having
global convergence. In the present work, we reverse the viewpoint.
Since the decoding procedure is performed with \textsc{lbp},
presumably the best encoding of the historical data is the one for
which \textsc{lbp}'s output is $\hat p_a$ in absence of ``real time''
information, that is when all the variables remain hidden ($\V^* = \emptyset$). 
This has actually been proposed
in~\cite{Wai06}, where it is proved in a specific case, that working
with the ``wrong'' model, i.e.\ the message passing approximate
version, yields better results from the decoding viewpoint. We will
come back on this later in Section~\ref{comparison}, when we will
compare various possible approximate models within this framework. 
In this paper, we propose  a new  approach, based on multiple fixed points
of \textsc{lbp} identification, able to deal both with the encoding and decoding 
procedure in a consistent way, suitable for real time applications.
The paper is organized as follows: our inference strategy is detailed in
Section~\ref{algorithmes}; in Section~\ref{mixture}, we specify the
problem to the inference of binary variables which distribution
follows a mixture of product forms and present some numerical results;
these are analyzed in Section~\ref{analysis} in the light of some
scaling limits where mean field equations become relevant, allowing
for a direct connection with the Hopfield model. In
Section~\ref{optimization} we propose a multi-parameter extension of
the model well suited to a continuous optimization, which allows to
enhance the performance of the model. Finally we conclude in Section~\ref{comparison}
by comparing our approach with other variant of \textsc{lbp} and giving
perspective for future developments. 

\section{LBP inference with marginal constraints}\label{algorithmes}

\subsection{The belief propagation algorithm}

The belief propagation algorithm~\cite{Pearl} is a message passing
procedure, with a joint probability measure like (\ref{eq:joint}) as input, and  
which output is a set of estimated marginal probabilities,
the beliefs $\bel_a(\x_a)$ (including single nodes beliefs
$\bel_i(x_i)$). The idea is to factor the marginal probability at a
given site as a product of contributions coming from neighboring
factor nodes, which are the messages. With our definition of the joint
probability measure, the updates rules read:
\begin{align}
m_{a\to i}(x_i) &\gets
\sum_{\x_{a\setminus i}} \psi_a(\x_a)\prod_{j\in a\setminus i} 
n_{j\to a }(x_j), \label{urules}\\[0.2cm]
n_{i \to a}(x_i) &\egaldef \phi_i(x_i)\prod_{a'\ni i, a'\ne a}
m_{a'\to i}(x_i), \label{urulesn}
\end{align}
where the notation $\sum_{\x_s}$ should be understood as summing all
the variables $x_i$, $i\in s\subset \V$, from $1$ to $q$. When the
algorithm converges, the resulting beliefs are
\begin{align}
\bel_i(x_i) &\egaldef 
\frac{1}{Z_i}\phi_i(x_i)\prod_{a\ni i} m_{a\to i}(x_i), \label{belief1}\\[0.2cm]
\bel_a(\x_a) &\egaldef 
\frac{1}{Z_a}\psi_a(\x_a)\prod_{i\in a} n_{i\to a}(x_i), \label{belief2} 
\end{align}
where $Z_i$ and $Z_a$ are the corresponding normalization constants
that make these beliefs sum to $1$. These constants reduce to $1$ when
$\G$ is a tree. In practice, the messages are normalized to have
\begin{equation}\label{eq:normalization}
 \sum_{x_i=1}^q m_{a\to i}(x_i)= 1.
\end{equation}

A simple computation shows that equations~(\ref{belief1}) and
(\ref{belief2}) are compatible, since (\ref{urules})--(\ref{urulesn})
imply that
\begin{equation}
\sum_{\x_{a\setminus i}} \bel_a(\x_a) = \bel_i(x_i).\label{eq:compat}
\end{equation} 

We can already address the \emph{inference} issue of the introduction:
inferring the law of all variables from the set $\V^*$ of variables on
which data is known is equivalent to evaluating the conditional
probability
\[
 \P(x_i|\x^*) = \frac{\P(x_i,\x^*)}{\P(\x^*)}.
\]
\textsc{lbp} is adapted to this case if a specific rule is defined
for known variables $i\in\V^*$: since the value of 
$x^*_i$ is known, there is no need to sum over possible values
and~(\ref{urulesn}) becomes
\begin{equation}\label{condurules}
n_{i \to a}(x_i) \egaldef 
\begin{cases}\phi_i(x_i)\prod_{a'\ni i, a'\ne a}m_{a'\to
  i}(x_i),&\text{if }i\notin \V^* \text{ or }x^*_i=x_i,\\
0,&\text{otherwise}.
\end{cases}
\end{equation}

\subsection{Setting the model with LBP}\label{heuristic}
Fixed points of \textsc{lbp} algorithm yield only approximate marginal probabilities
of $\P(\x)$ when all the functions $\psi_a$ and $\phi_i$ are known and considered as
an input.
Conversely, assume that a set of marginal distributions $\{\hat p\}$
is given such that, for all $a\in\F$ and $i\in a$,
\[
  \sum_{\x_{a\setminus i}}\hat p_a(\x_a)=\hat p_i(x_i)
  \quad\text{and}\quad
  \sum_{\x_i}\hat p_i(x_i)=1.
\]
Finding the set of $\{\psi_a\}$ and $\{\phi_i\}$ such that 
the marginals of  the joint probability~(\ref{eq:joint})
match $\{\hat p\}$ is a difficult inverse problem. Instead if we impose that the
\emph{approximation via \textsc{lbp}} of these marginals matches $\{\hat p\}$,
we face a much simpler problem:
owing to its \emph{reparametrization property}~\cite{Wain}, \textsc{lbp} can
provide good candidates for $\psi_a$ and $\phi_i$ that admit a fixed
point where $\bel_a(\x_a) = \hat p_a(\x_a)$, $\forall a\in\F$, and
therefore $\bel_i(x_i) = \hat p_i(x_i)$, $\forall i\in\V$.

We look for a fixed point that satisfies
(\ref{urules})--(\ref{urulesn}) in addition to this constraint.
Normalization constants introduced in (\ref{belief1})--(\ref{belief2})
play no role in the present discussion so we ignore them here.
Using (\ref{belief1})--(\ref{belief2}) to rewrite (\ref{eq:joint}),
one sees that the knowledge of one set of beliefs is sufficient to
determine the underlying \textsc{mrf} uniquely:
\begin{align*}
\P(\x) 
&= \prod_{i\in\V}\phi_i(x_i) \prod_{a\in\F}\psi_a(\x_a)
=\prod_{i\in\V}\bel_i(x_i)
 \prod_{a\in \F}\frac{\bel_a(\x_a)} {\prod_{i\in a} \bel_i(x_i)}.
\end{align*}
It is therefore tempting to choose the functions appearing
in~(\ref{eq:joint}) as follows.
\begin{equation}
  \hat\phi_i(x_i)  \egaldef \hat p_i(x_i), \qquad
  \hat\psi_a(\x_a) \egaldef\frac{\hat p_a(\x_a)}{\prod_{i\in a}\hat p_i(x_i)}.
\label{bethephipsi}
\end{equation}

This leads to the following formulation for the BP algorithm
\begin{equation}
m_{a\to i}(x_i) \gets \sum_{\x_{a\setminus i}} \frac{\hat p_a(\x_a)}{\hat p_i(x_i)}
\biggl[\prod_{j\in a\setminus i}
\prod_{a'\ni j, a'\ne a}m_{a'\to j}(x_j)\biggr],\label{urulebethe}
\end{equation}
which obviously admits $m_{a\to i}(x_i)\equiv 1$ as a fixed point, and
leads to the beliefs 
\begin{equation}
\bel(\x_a) = \hat p(\x_a)\quad\forall a\in\F
\text{ and }\ 
\bel(x_i) = \hat p(x_i)\quad\forall i\in\V\label{goodbelief},
\end{equation}

This choice of functions~(\ref{bethephipsi}) may seem arbitrary at
first sight. It has however already been proposed in~\cite{Wai06} and,
in a slightly different problem of ML estimation, in~\cite{Wain2}.
Moreover, the following proposition shows that any other choice of
$\psi$ and $\phi$ is actually equivalent:
\begin{prop}\label{proppsi}
Any given set of functions $\psi$ and $\phi$ such that \textsc{lbp} yields the
prescribed fixed point~(\ref{goodbelief}), provides exactly the same
set of fixed points, including their stability properties, as
$\hat\psi$ and $\hat\phi$ would.
\end{prop}
\begin{proof}
Assume that there exists a set of messages $m^0$ which is a fixed
point of \textsc{lbp} and such that
\begin{align*}
\hat p_a(\x_a) &\egaldef \psi_a(\x_a)\prod_{i\in a} \Bigl[\phi_i(x_i)\prod_{a'\ni i, a'\ne a}
m^0_{a'\to i}(x_i)\Bigr],\\
\hat p_i(x_i) &\egaldef \phi_i(x_i)\prod_{a\ni i} m^0_{a\to i}(x_i).
\end{align*}
Then it is possible to express $\phi$ and $\psi$ as
\begin{align*}
\phi_i(x_i)
 = \frac{\hat \phi_i(x_i)}
         {\prod_{a\ni i}m^0_{a\to i}(x_i)},\qquad
\psi_a(\x_a) 
= \hat\psi_a(\x_a)\prod_{j\in a} m^0_{a\to j}(x_j),
\end{align*}
and relations~(\ref{urules})--(\ref{urulesn}) rewrite
\begin{align*}
\frac{m_{a\to i}(x_i)}{m^0_{a\to i}(x_i)} &\leftarrow 
\sum_{\x_{a\setminus i}} \hat\psi_a(\x_a)\prod_{j\in a\setminus i} 
n_{j\to a }(x_j)m^0_{a\to j}(x_j),\\[0.2cm]
n_{i \to a}(x_i) &= \frac{\hat\phi_i(x_i)}{m^0_{a\to j}(x_j)}\prod_{a'\ni i, a'\ne a}
\frac{m_{a'\to i}(x_i)}{m^0_{a'\to i}(x_i)},
\end{align*}
Therefore, $m_{a\to i}(x_i)/m^0_{a\to i}(x_i)$ stands for the set of
fixed point messages that would have been obtained with functions
$\hat\psi$ and $\hat\phi$, and the two versions of the algorithm are
equivalent.
\end{proof}

\subsection{Controlling the strength of the interaction}\label{strengthpar}
The structure of the factor graph on which \textsc{lbp} is supposed to
be run is more or less imposed by the data. For example, if mutual
information is given for each pair of variables, we then have a
complete pairwise factor graph. In that case, \textsc{lbp}, which is
well adapted to sparse graphs, will overestimate the mutual
information between variables. To overcome this flaw, we introduce a
single real parameter $\alpha>0$, to be roughly interpreted as an
inverse temperature, which purpose is to moderate (or possibly
amplify) the interaction between variables when the connectivity gets
large. This is done through a geometric mean with the independent case,
by replacing $\hat p_{a}$ with $\hat p_{a}^\alpha(\prod_{i\in a}\hat
p_i)^{(1-\alpha)}$. The model (\ref{bethephipsi}) is then rewritten as
\begin{equation}\label{def:lbpmeasure}
\phi_i(x_i) 
  \egaldef \hat p_i(x_i), \qquad
\psi_{a}(\x_a) 
   \egaldef\Bigl(\frac{\hat p_{a}(\x_a)}
                      {\prod_{i\in a}\hat p_i(x_i)}\Bigr)^\alpha.
\end{equation}
This definition allows to interpolate between a situation with strong
interaction ($\alpha\gg 1$) and a situation with weak interactions
($\alpha\simeq 0$). Note that for $\alpha\ne 1$, $\hat p$ is not
anymore a predefined fixed point of the \textsc{lbp} scheme. However,
Section~\ref{mixture} will show that (\ref{def:lbpmeasure}) does yield
consistent results. In fact a quite similar deformation of the model
has been proposed in~\cite{WiHe}, which we discuss later in
Section~\ref{comparison}.

A related approach would have been to replace $\hat p_a$ with
$\beta\hat p_a+(1-\beta)\prod_{i\in a}\hat p_i$; this would preserve the
single variables beliefs, without however affecting the results we
present in a sensible way. Note that this is actually equivalent to
replacing $\hat\psi_a$ by $\beta\hat\psi_a+(1-\beta)$.

Finally, an optimization with respect to
the graph structure could be done afterwards, but we won't explore
this possibility in the present work. Instead we will focus in
Section~\ref{optimization} on the possibility to associate various
parameter values to different types of edges, and to perform an
optimization procedure with respect to these parameters.

\section{Inferring a hidden mixture of product forms}\label{mixture}
\providecommand{\Pref}{\P_{\mathrm{ref}}}
\subsection{Experimental setting}\label{experiment}
To test the ideas developed in the previous section, we assume a
hidden mixture model on a set $\V$ of variables with cardinality $N$
of the form
\begin{equation}\label{def:mixture}
\Pref(\x)\egaldef \frac{1}{C}\sum_{c=1}^{C}\prod_{i\in\V}p_i^c(x_i),
\end{equation}
where $\x=\{x_i, i\in\V\}$ is a sequence of binary variables
($x_i\in\{0,1\}$), $C$ is the number of components of the mixture
which are superimposed, and $p_i^c(\cdot)$ is the single site marginal
corresponding to variable $i$ for model $c$.  The main virtue of this
simplified testbed is that the performance of the approach we propose
can be easily compared with theoretical bounds.

In order to apply our inference method, we assume that the
distribution (\ref{def:mixture}) is unknown as well as the number $C$
itself. The input of the algorithm is the set of $1$- and
$2$-variables frequency statistics $\hat p_i(x_i)$ and $\hat
p_{ij}(x_i,x_j)$. Part of the freedom in choosing a \textsc{lbp} model is in
the graph design. While the available data dictates a pairwise factor
graph (each factor node is connected at most to two variables), it is
still possible to choose which pairs of variables will be connected.
To this end, we apply a simple pruning procedure, by selecting the
links $(i,j)$ for which the quantity (to be interpreted in
Section~\ref{analysis})
\[
\Bigl\lvert\log \frac{\hat p_{ij}(1,1)\hat p_{ij}(0,0)}
{\hat p_{ij}(0,1)\hat p_{ij}(1,0)}\Bigr\rvert\ge \epsilon,
\]  
where $\epsilon$ is some positive threshold. We denote by $K$ the mean
connectivity of the resulting graph.

Although (\ref{def:mixture}) is quite general, the tests are conducted
with $C \ll 2^N$, in the limit were the optimal sequences
$\x^{c,\mathrm{opt}}$ of each component $c$ (i.e.\ with highest
probability weight in the restricted distribution) have mutual Hamming
distance of order $N/2$. The single sites probabilities $p_i^c =
p_i^c(1)$, corresponding to each component $c$, are generated randomly
as i.i.d.\ variables, 
\[
p_i^c = \frac{1}{2}(1+\tanh h_i^c)
\]
with $h_i^c$ uniformly distributed in some fixed interval $[-h_{max},+h_{max}]$.
The mean of $p_i^c$ is therefore $1/2$ and its variance reads
\[
v \egaldef \frac{1}{4}{\mathbb E}_h\bigr(\tanh^2(h)\bigl)\ \in[0,1/4].
\]
This parameter $v$ implicitly fixed by $h_{max}$
fixes the average level of
``polarizability'' of the variables in each cluster: $v=0$ corresponds
to $p_i^c=1/2$ while $v=1/4$ corresponds to $p_i^c\in\{0,1\}$ with
equal probability. The optimal configuration for each
component is given by
\[
x_i^{c,\mathrm{opt}} = \ind{p_i^c>0.5}.
\]
After fixing $N$ and $C$, we randomly generate a set $\{p_i^c,i\in
\V,1\le c\le C\}$ for a given value of $v$. The pruning of the graph
is performed to reach a prescribed average connectivity $K$. 
Then two types of experiments are performed:
\begin{itemize}
\item \textbf{BP fixed points search}, with the help of an evanescent
guiding field $h_t\to_{t\to\infty}~0$: if $t$ is the iteration step,
we bias the \textsc{lbp} updates (\ref{urulesn}) in the direction of
one of the patterns by replacing $\phi_i(x_i)$ by
\[
\phi_i^t(x_i) = \phi_i(x_i)e^{h_t(2x_i-1)(2x_i^c-1)},
\]
so that if there is a belief propagation fixed point correlated to the
pattern ${\bf p}^c$, the field $h_t$, which decays geometrically,
helps to find the corresponding attractor. The corresponding set of
beliefs $\>\bel^c$ which is obtained is then compared to ${\bf p}^c$.
\item \textbf{decimation}: Sequences $\x^c$ are sampled for each component
$c$ of (\ref{def:mixture}), and the decoding algorithm is tested
successively (with no guiding field) after gradually revealing the
elements of the sequence in a random order, and $\rho$ denotes the
fraction of observed variables. To each $\x^c$ and $\rho$, the output
is again a set of beliefs $\>\bel^c$ for the hidden variables to be
compared with the exact conditional marginals extracted from
(\ref{def:mixture}).
\end{itemize}
The following indicators are used to assess the prediction success
rate ($R$), the belief error ($E$) and the Kullback-Leibler error
($D_{\mathrm{KL}}$) of the algorithm when the values $\{x^c_i,\ i\in\V^*\}$ are
known
\begin{align*}
R &\egaldef\frac{1}{C}\frac{1}{|\V\setminus\V^*|}
          \sum_{c=1}^C\sum_{i\in\V\setminus\V^*}
\ind{\bel_i^c(1)>0.5} x^c_i+ \ind{\bel_i^c(1)\le0.5}(1-x^c_i),\\
E &\egaldef \frac{1}{C}\frac{1}{|\V\setminus\V^*|}
          \sum_{c=1}^C\sum_{i\in\V\setminus\V^*} \sum_{x\in\{0,1\}} 
\Bigl\lvert \bel_i^c(x) - \Pref(x_i=x|\x^c_{\V^*})\Bigr\rvert,\\
D_{\mathrm{KL}} &\egaldef \frac{1}{C}\frac{1}{|\V\setminus\V^*|} 
          \sum_{c=1}^C\sum_{i\in\V\setminus\V^*} \sum_{x\in\{0,1\}}  
\bel_i^c(x)\log\frac{\bel_i^c(x)}{\Pref(x_i=x|\x^c_{\V^*})}.
\end{align*}
where $\Pref(x_i|\x^*)$ is the conditional distribution of $x_i$ once
a certain number of variables $\x^*$ have been fixed, computed exactly
from the hidden model (\ref{def:mixture}). $R$ is to be compared with
the following expected success rate, which would be obtained by making
use of the hidden underlying model,
\[
R^{(0)} \egaldef\frac{1}{C}\frac{1}{|\V\setminus\V^*|}
          \sum_{c=1}^C\sum_{i\in\V\setminus\V^*}
\ind{\Pref(x_i|\x^c_{\V^*})>0.5} x^c_i+ \ind{\Pref(x_i|\x^c_{\V^*})\le0.5}(1-x^c_i).
\]

\subsection{Preliminary Observations}\label{results}
\begin{figure}
\centering
\includegraphics*[width=0.49\linewidth]{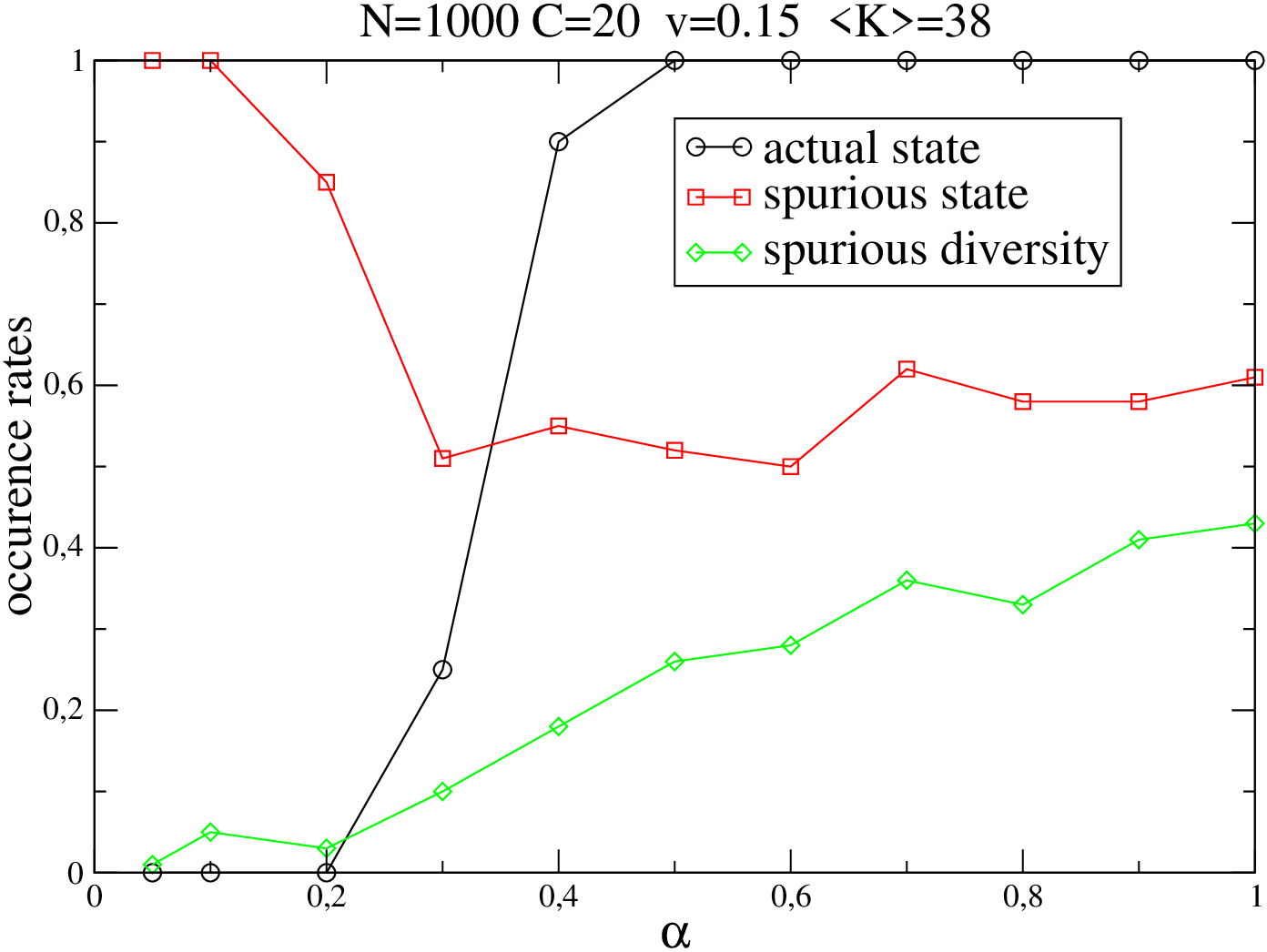}
\hfill
\includegraphics*[width=0.49\linewidth]{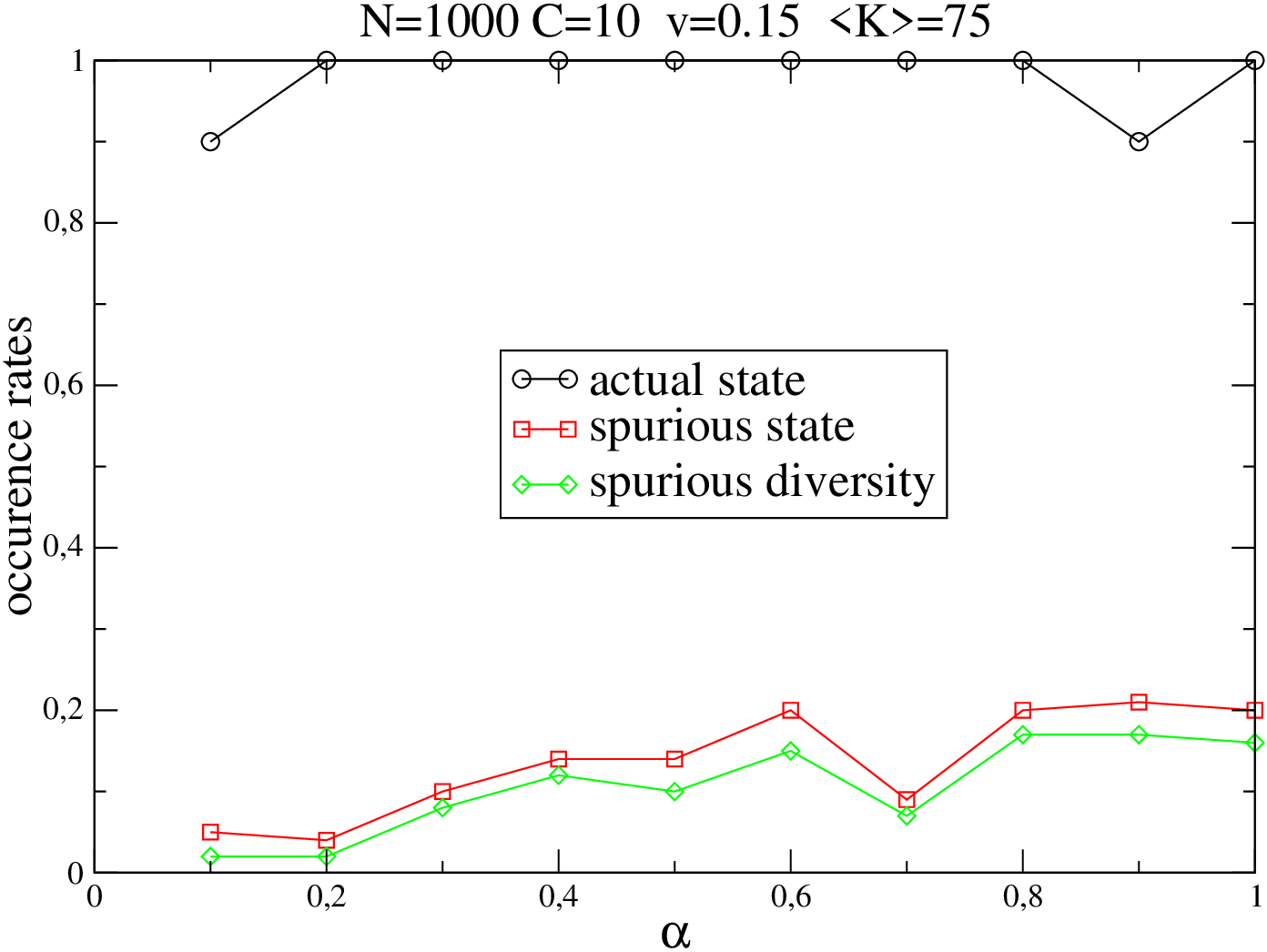}
\caption{\label{spurious} Proportion of actual fixed points
  (circles) found by \textsc{lbp}, probability of convergence toward a spurious fixed
  point (squares) from a random initialization, and number of different spurious fixed points
  divided by the number of runs (100) (left: $C/K=0.52$, right $C/K=0.13$).}
\end{figure}

To assess this approach, we look first at the quality of the encoding
(Figure~\ref{spurious}), by studying the nature of the fixed points
when all the variables are hidden. Regarding to the quality of the
encoding, we check whether the fixed points of \textsc{lbp} correctly represent
the component of the probability mixture, by guiding \textsc{lbp} at the
beginning of the iterations.  In a second step, we run \textsc{lbp} without
guiding, and measure the probability to converge to a spurious fixed
point and the diversity of these fixed points.  We observe that there
is a specific ratio $\eta^*$ of $\eta= C/K$, below which it is always
possible to find a value of $\alpha$ such that a fixed point is
associated to each encoded state and no other spurious fixed point is
present.  In that case, as $\alpha$ varies, 3 different regimes are to
be found: when $\alpha$ is too small, only one fixed point is present,
in the intermediate range of $\alpha$ of interest all fixed points
correspond to the encoded states, and for larger $\alpha$, a
proliferation of fixed points occurs, while the ones corresponding to
the encoded states are destabilized. This will be analyzed in
Section~\ref{analysis}.
\begin{figure}
\centering
\includegraphics*[width=0.49\linewidth]{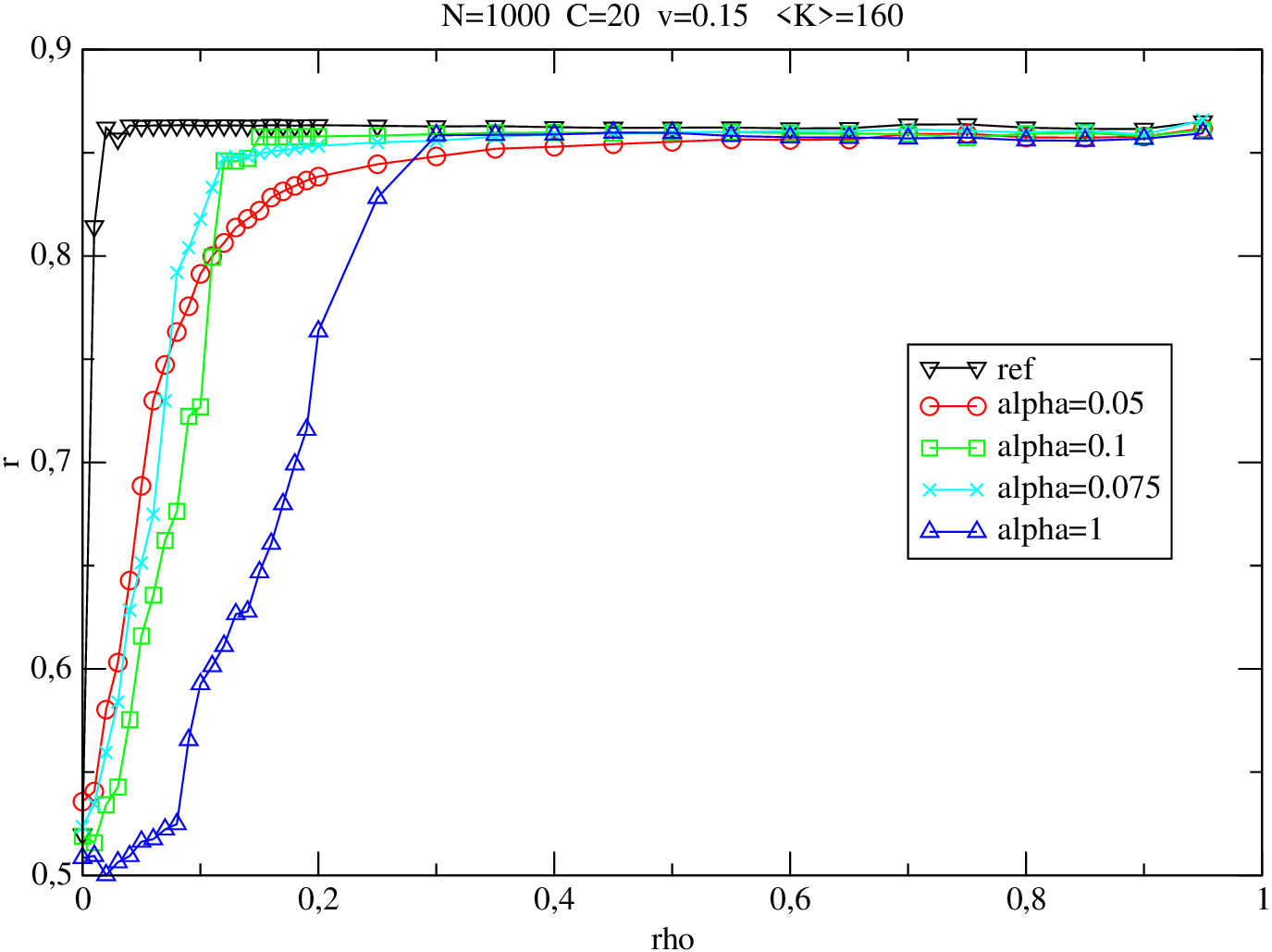}
\hfill
\includegraphics*[width=0.49\linewidth]{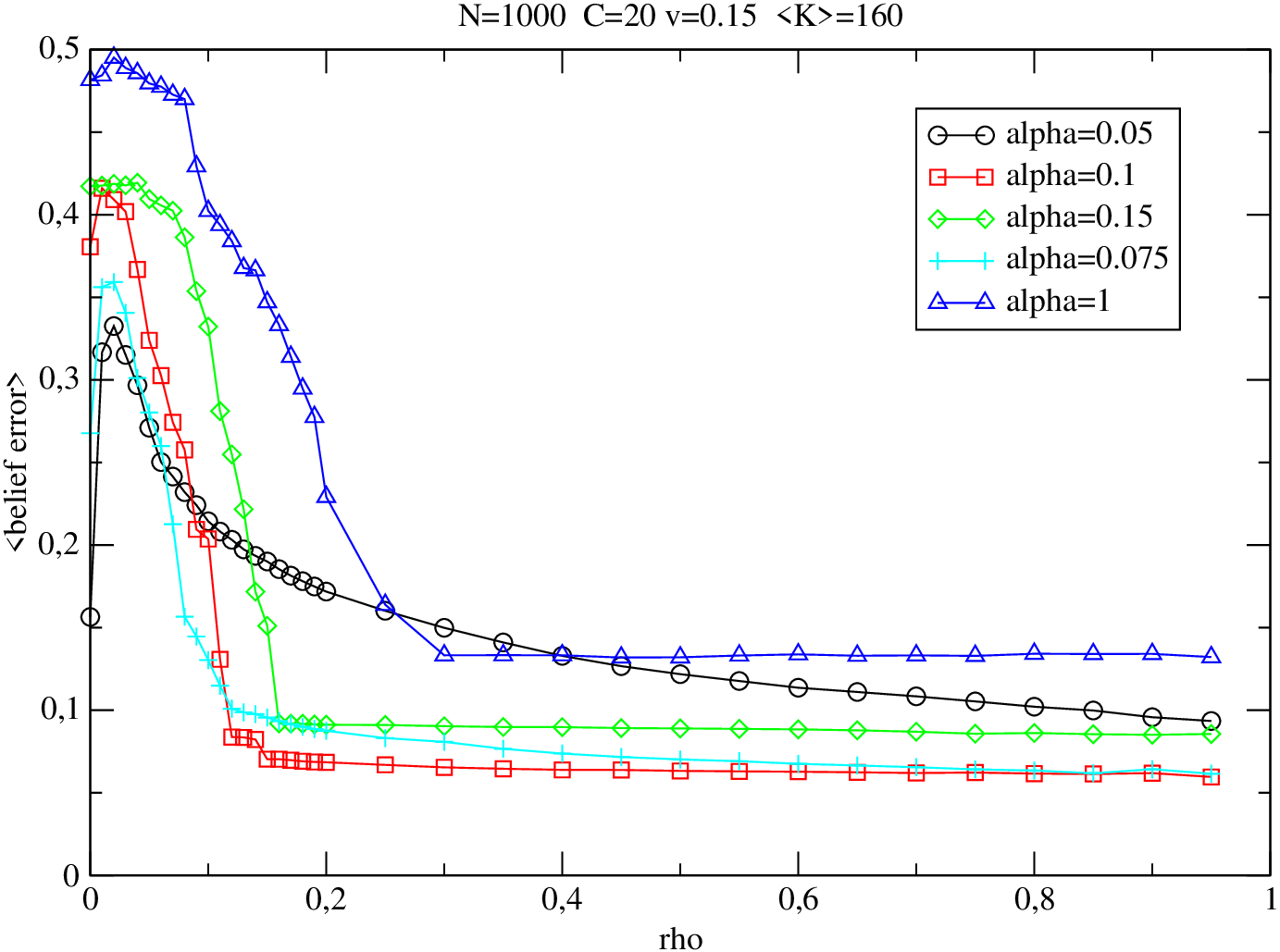}
\caption{\label{alpha} Influence of $\alpha$ on the inference success
  rate  $R$ (left) and on the belief error $E$ (right) for a fixed ratio 
$C/K=0.125$.}
\end{figure}
\begin{figure}
\centering
\includegraphics*[width=0.49\linewidth]{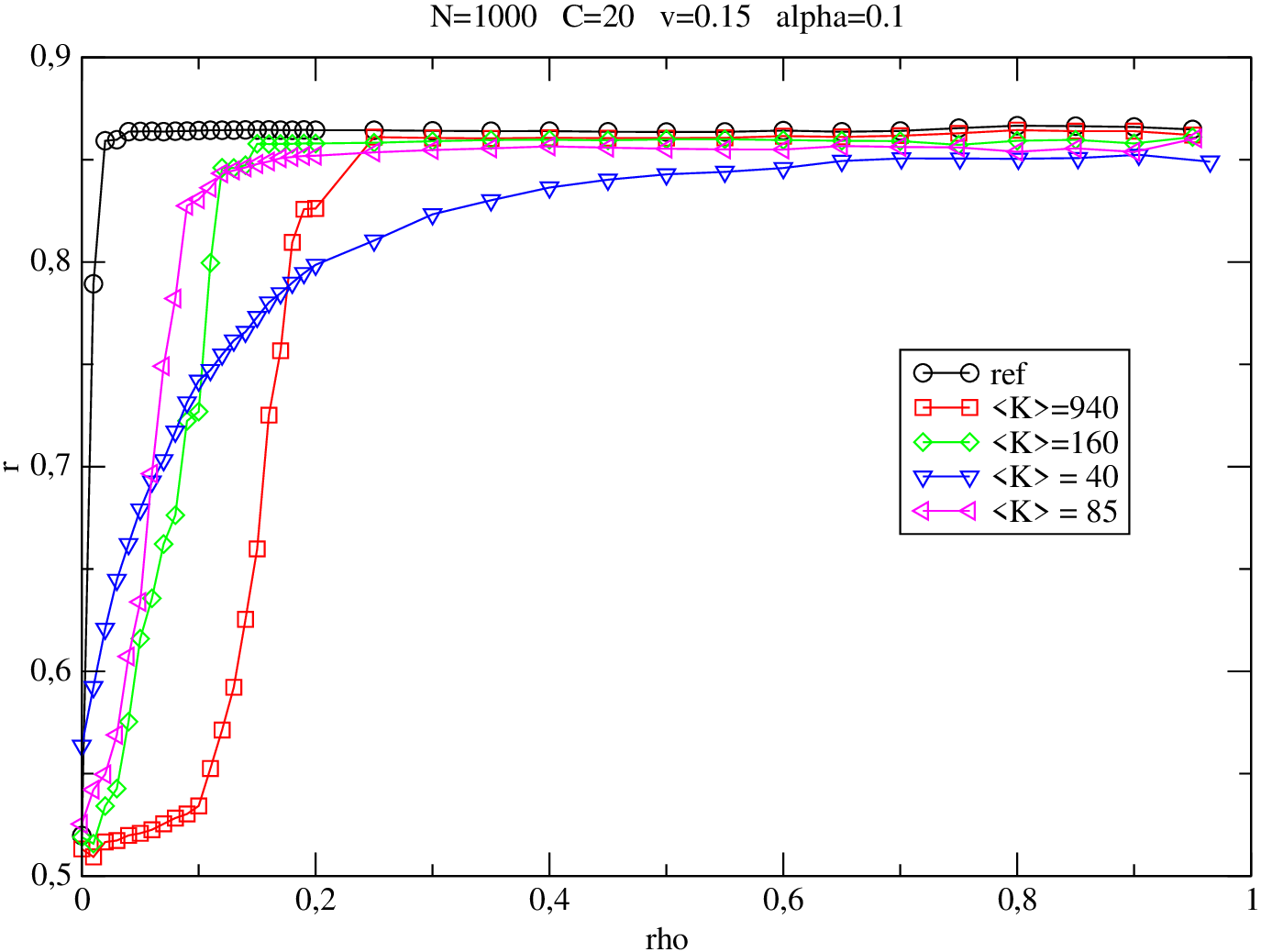}
\hfill
\includegraphics*[width=0.49\linewidth]{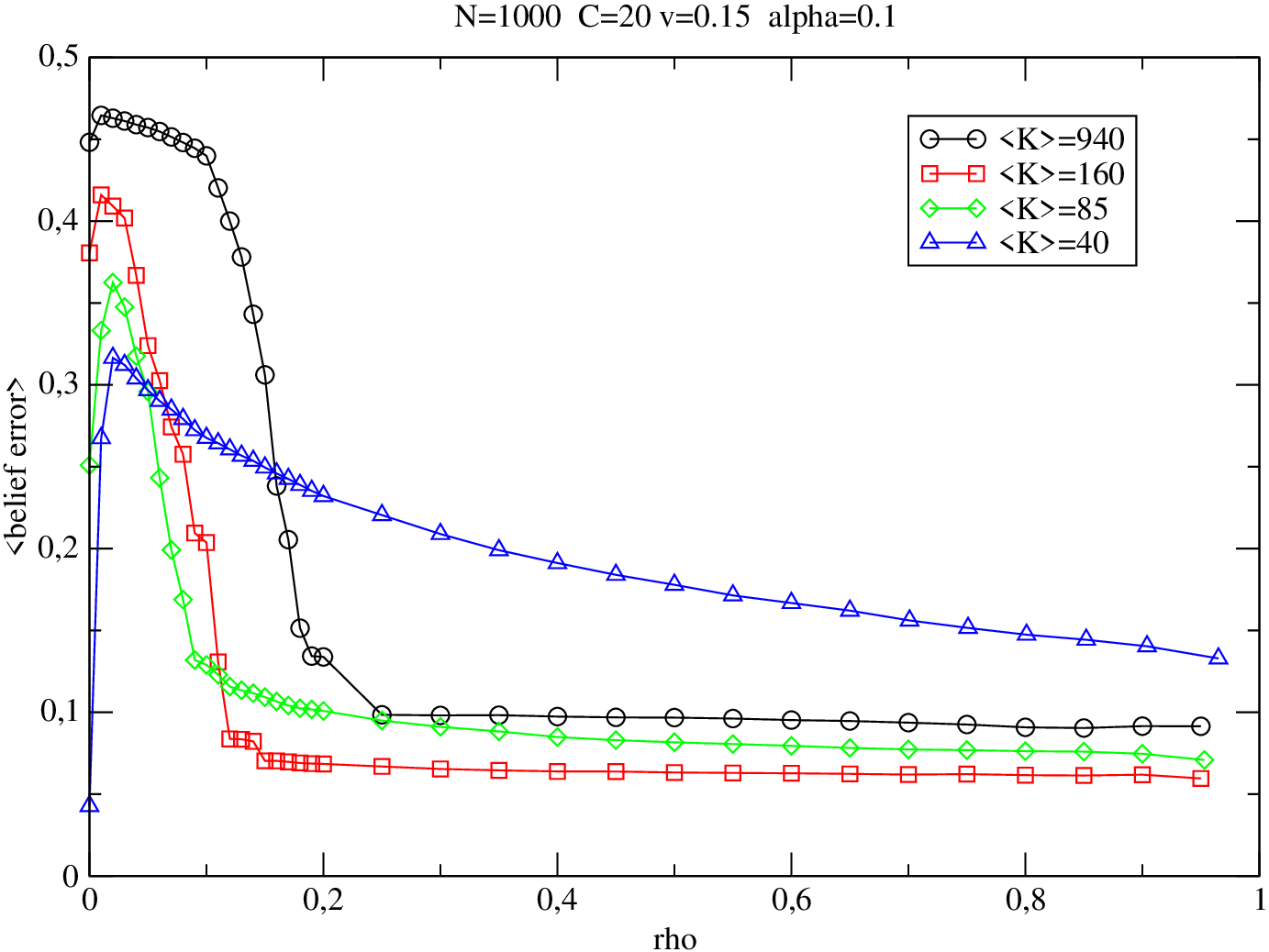}
\caption{\label{Km} Influence of the pruning on the inference success
  rate $R$ (left) and on the belief error $E$ (right) at given $\alpha$.}
\end{figure}

The second point is the efficiency and reliability of the decoding
procedure. The question is to measure how well \textsc{lbp} performs (in term
of $R$ and $E$ defined in previous section) when the proportion of
known variables increases. Figure~\ref{alpha} shows, for several values
of $\alpha$, the evolution of $R$ and $E$ as the
proportion $\rho$ of revealed variables increases. This is compared with
the ideal reconstruction rate $R^{(0)}$, which would be obtained
from the  underlying mixture model. Typically, for the
optimal value of $\alpha$, knowing $10\%$ of the variables is
sufficient to reach the optimal inference rate (see left plot). When
looking at the mean absolute value error on the beliefs $E$, an error
of less than $0.1$ is generally achieved with this optimal choice of
$\alpha$ (see right plot). The effect of the pruning procedure is
shown in Figure~\ref{Km}. The performance deteriorates smoothly, when
the parameter $v$ decreases. 

\section{Mean-Field analysis}\label{analysis}
\subsection{Connection with the Hopfield model for large $C$}
The connection between the \textsc{lbp} algorithm and statistical
physics has been recognized recently.  It has been established that
the \textsc{lbp} fixed points correspond to local minima of the Bethe
free Energy~\cite{YeFrWe}, and that the \textsc{lbp} scheme is
actually providing solutions to the mean field \textsc{tap}
equations~\cite{KaSa}. Let us consider the asymptotic situation corresponding to having both
$C$ and $K$ are large. Using spin variables of statistical physics
$s_i=2x_i-1$, the measure (\ref{def:lbpmeasure}) may be cast in the
standard form of the disordered Ising model
\[
\P(\>s) = \frac{1}{Z}e^{-\beta H[\>s]},
\]
with $\beta$ the inverse temperature (which is arbitrary for the moment) and
the definition
\[
H[\>s] \egaldef -\frac{1}{2}\sum_{i,j}J_{ij}s_i s_j-\sum_i h_i s_i.
\]
The identification with the marginals gives:
\begin{align*}
\beta J_{ij} &= \frac{\alpha}{4}\log\frac{\hat p_{ij}(1,1)\hat
  p_{ij}(0,0)}{\hat p_{ij}(0,1)\hat p_{ij}(1,0)},\\[0.2cm]
\beta h_i &= \frac{1-\alpha K_i}{2}\log\frac{\hat p_i(1)}{\hat p_i(0)}
+\frac{\alpha}{4}\sum_{j\in i}\log\frac{\hat p_{ij}(1,1)\hat
  p_{ij}(1,0)}{\hat p_{ij}(0,1)\hat p_{ij}(0,0)},
\end{align*}
with
\begin{align*}
\hat p_i(\tau) &\egaldef \frac{1}{2C}\sum_{c=1}^C \bigl(1+(2\tau-1)(2p_i^c-1)\bigr),\\ 
\hat p_{ij}(\tau,\tau') &\egaldef \frac{1}{4C}\sum_{c=1}^C 
\bigl(1+(2\tau-1)(2p_i^c-1)\bigr)\bigl(1+(2\tau'-1)(2p_j^c-1)\bigr).
\end{align*}
for $\tau$ and $\tau'$ in $\{0,1\}$.
Let 
\begin{equation}\label{def:xi}
\xi_i^c \egaldef \frac{p_i^c(1)-\frac{1}{2}}{\sqrt{v}}\qquad \xi_i\egaldef \frac{1}{C}\sum_{c=1}^C \xi_i^c
\qquad \xi_{ij}\egaldef \frac{1}{C}\sum_{c=1}^C \xi_i^c\xi_j^c - \xi_i\xi_j.
\end{equation}
For large $C$, we have, in distribution
\begin{equation}\label{eq:normal}
\lim_{C\to\infty}{\sqrt C}\xi_i \sim \mathcal{N}(0,1),\qquad 
\lim_{C\to\infty}{\sqrt C}\xi_{ij} \sim \mathcal{N}(0,1).
\end{equation}
where $\mathcal{N}(0,1)$ denotes a normal variable with unit variance.
Using this notation, and assuming $C\gg1$, we have
\begin{align}
\beta J_{ij} &= 4\alpha v \xi_{ij} + O\bigl(\frac{1}{C^{3/2}}\bigr)\label{eq:limcoup1}\\[0.2cm]
\beta h_i &= 2\sqrt{v}\xi_i - 8\alpha v^{3/2}\sum_{j\in i}\xi_j\xi_{ij} + 
K O\bigl(\frac{1}{C^{3/2}}\bigr)\label{eq:limcoup2}.
\end{align}
for fixed connectivity $K$.
Note that, in addition to (\ref{eq:normal}), we have 
\[
\lim_{\substack{C\to\infty\\ K\to\infty}} \frac{C}{\sqrt K}\sum_{\substack{j=1\\j\ne i}}^K\xi_j\xi_{ij} \sim \mathcal{N}(0,1),
\]
and that the two terms present in $h_i$ are uncorrelated at first
order (the covariance between $\xi_i$ and $\xi_{ij}\xi_j$ is zero). In
this form, the Hamiltonian is similar to the one governing the
dynamics of the Hopfield neural network model~\cite{Hopfield,MePaVi}.
Considering the canonical form of the Hamiltonian chosen
in~\cite{AmGuSo3},
\[
H[\>s] = -\frac{1}{2N}\sum_{i,j,c}s_i\xi_i^cu_i^{(K)}u_j^{(K)}\xi_j^c s_j -\sum_{i,c}h_i^c\xi_i^c s_i,
\]
adapted to a non-complete graph,
the inverse temperature then reads
\[
\beta = \frac{4\alpha v K}{C}
\]
and 
\[
h_i^c = \frac{C}{2\alpha K\sqrt v} -  \frac{2C\sqrt v}{K}\sum_{j\in i} \xi_{ij}.
\]
The coefficients $u_i^{(K)}$ are the components of the Perron vector
normalized to $\sqrt N$ (so that $u_i^{(K)} = O(1)$), associated to
the largest eigenvalue $K$ of the incidence matrix\footnote{Here we
  keep track of the fact that we possibly deal with a non-complete graph with
  arbitrary topology given by some incidence matrix $A$: to each edge
  $(ij)$ preserved by the pruning procedure is associated the element
  $a_{ij}=1$, while other elements are set to $0$.  Under the
  hypothesis that the second eigenvalue is sub-dominant w.r.t. $K$ (it
  is generally the case when for example the connectivity is extensive
  with the size of the system), only the Perron eigenvector is to be
  considered in the mean field theory.}. When the graph has some
permutation symmetry with a uniform connectivity, $u_i^{(K)}$ reduces to $1$
and $K$ to this connectivity. $K$ is considered from now on as an extensive
parameter.

\subsection{Phase diagram}
The mean-field theory of the Hopfield model has been solved by Amit,
Gutfreund and Sompolinsky in~\cite{AmGuSo3} using replica's
techniques, results which were soon confirmed with help of the cavity
method~\cite{MePaVi}, and put later on even firmer mathematical
grounds in~\cite{Talagrand}.  In this section we can simply read off some
properties of our model from this mean field theory. The order
parameter introduced by~\cite{AmGuSo3} is
\begin{equation}\label{def:op}
\mu_c \egaldef  \frac{1}{N}
\sum_{i=1}^N{\mathbb E}_{S,\xi}\bigl(u_i^{(K)}\xi_i^c s_i\bigr),\qquad\forall c=1,\ldots,C.
\end{equation}
where the expectation comprises both thermal averages and expectation
with respect to the quenched disorder variables $\xi_i^c$.  This
quantity measures the correlation between the spin bias in each
components with the local magnetization. The projection on an arbitrary Perron vector
has been taken into account for sake of generality. 
Two cases are at stake in the thermodynamic limits.

\paragraph{(i) $C$ is large but fixed when $N\to \infty$.} 
In that case, considering that
\[
\beta \egaldef \frac{4v}{C}\lim_{N\to\infty} \alpha(N) K(N),
\]
is a fixed parameter in the thermodynamic limit, then
the mean-field free energy per variable directly adapted from~\cite{AmGuSo3} reads,
\[
f^{(N)}[\vec\mu,\vec\xi] \egaldef \frac{\beta}{2}\sum_c \mu_c^2 - 
\frac{1}{N}\sum_i\log\Bigl[2\cosh\bigl(\beta
\sum_{c}u_i^{(K)}(\xi_i^c-\xi_i)\mu_c -2\sqrt v \xi_i\bigr)\Bigr],
\]
where subdominant terms in the $1/C$ expansion are implicitly neglected. 
The stable thermodynamical states are then obtained by solving the
saddle point equation, which reads
\begin{align*}
\mu_c &= \frac{1}{N}\sum_{i=1}^N u_i^{(K)}(\xi_i^c-\xi_i)
\tanh\bigl(\beta\sum_{c}u_i^{(K)}(\xi_i^c-\xi_i)\mu_c -2\sqrt v \xi_i\bigr),\\[0.2cm]
&= \frac{1}{N}\sum_{i=1}^N u_i^{(K)}(\xi_i^c-\xi_i)
\tanh\bigl(\beta\sum_{c}\xi_i^c(u_i^{(K)}\mu_c-\bar\mu)\bigr),\qquad\forall c=1,\ldots,C.
\end{align*}
The last line is obtained after using that from the first equation
$\vec\mu$ is transverse and after defining
\[
\bar\mu \egaldef \frac{2\sqrt v}{C\beta}.
\]
These equations are very similar to the one obtained in~\cite{AmGuSo1}
and so are their solutions. For $\beta>\beta_c=1$, $2C$
thermodynamically stable states, referred to as Marris-states
in~\cite{AmGuSo1}, appear. Each one of these states is macroscopically
correlated or anti-correlated to one of the mixture component, i.e.\ a
single component $\mu_c$ acquires a finite value. They are the only
stable states up to some threshold value of $\beta$, where
mixed stable states do appear.

\paragraph{(ii) The number of components is extensive: $C = \eta K$.}
 
\begin{figure}[ht]
\centering
\includegraphics*[width=0.49\linewidth]{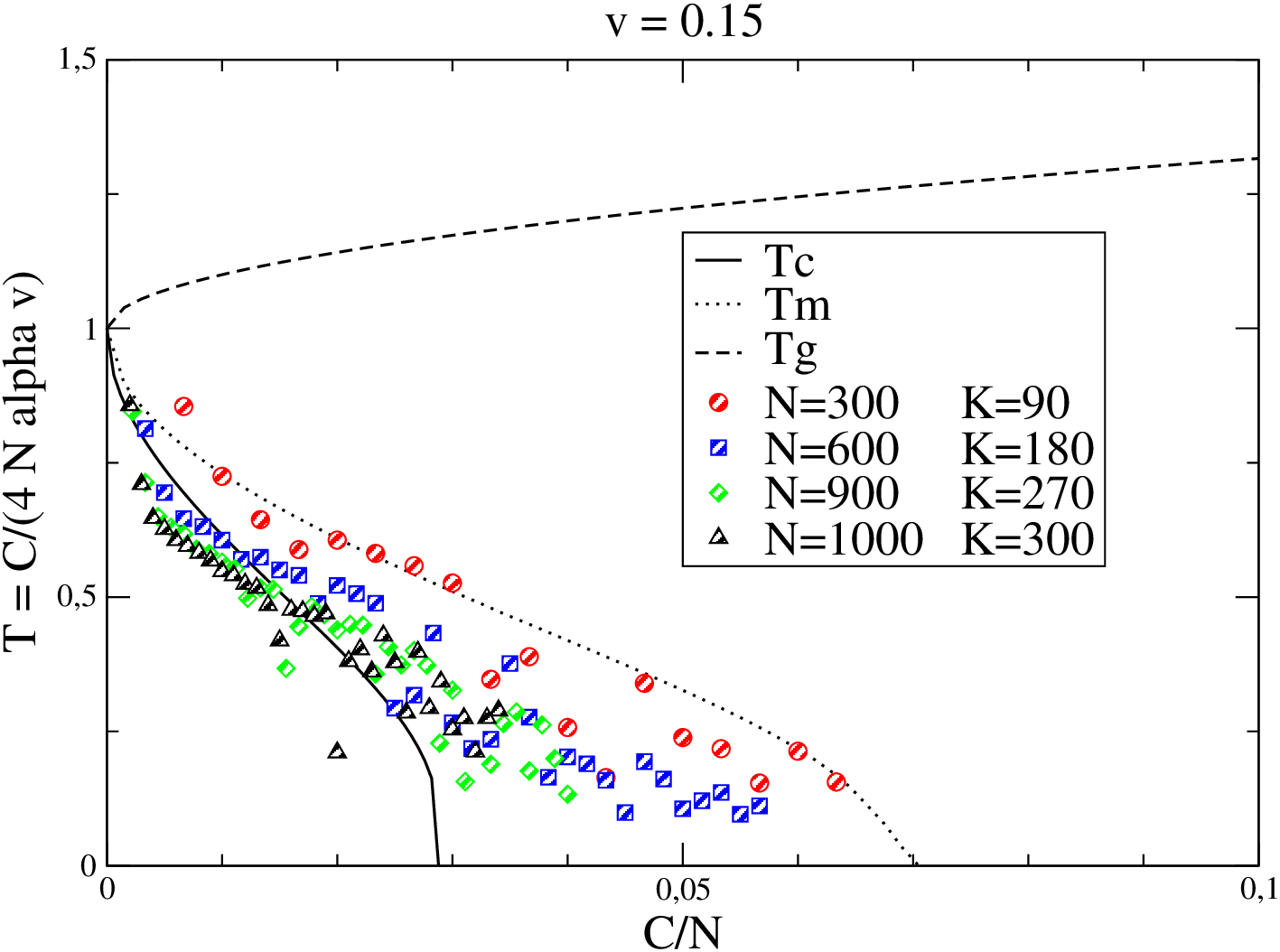}\hfill
\includegraphics*[width=0.47\linewidth]{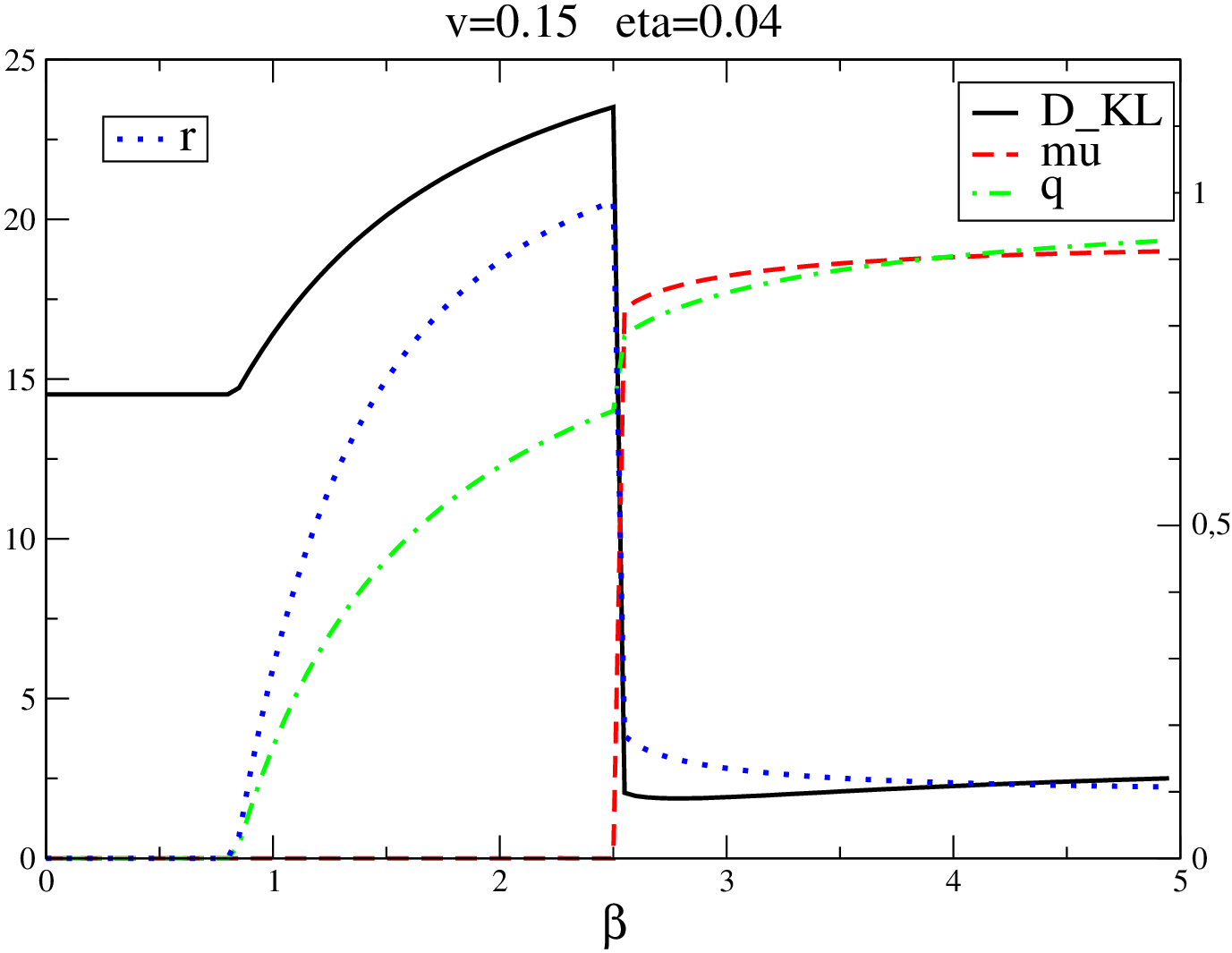}

\includegraphics*[width=0.49\linewidth]{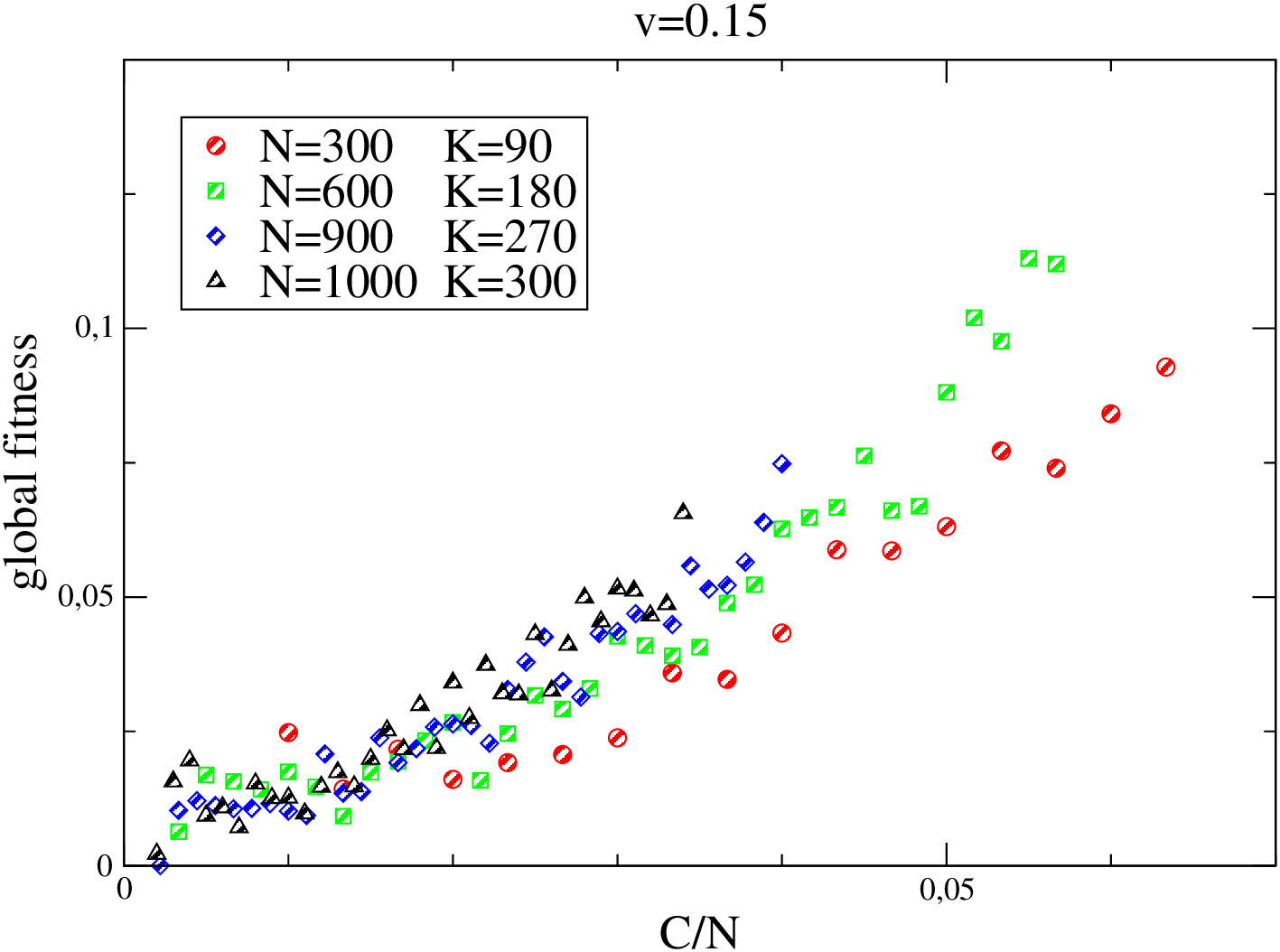}\hfill
\includegraphics*[width=0.49\linewidth]{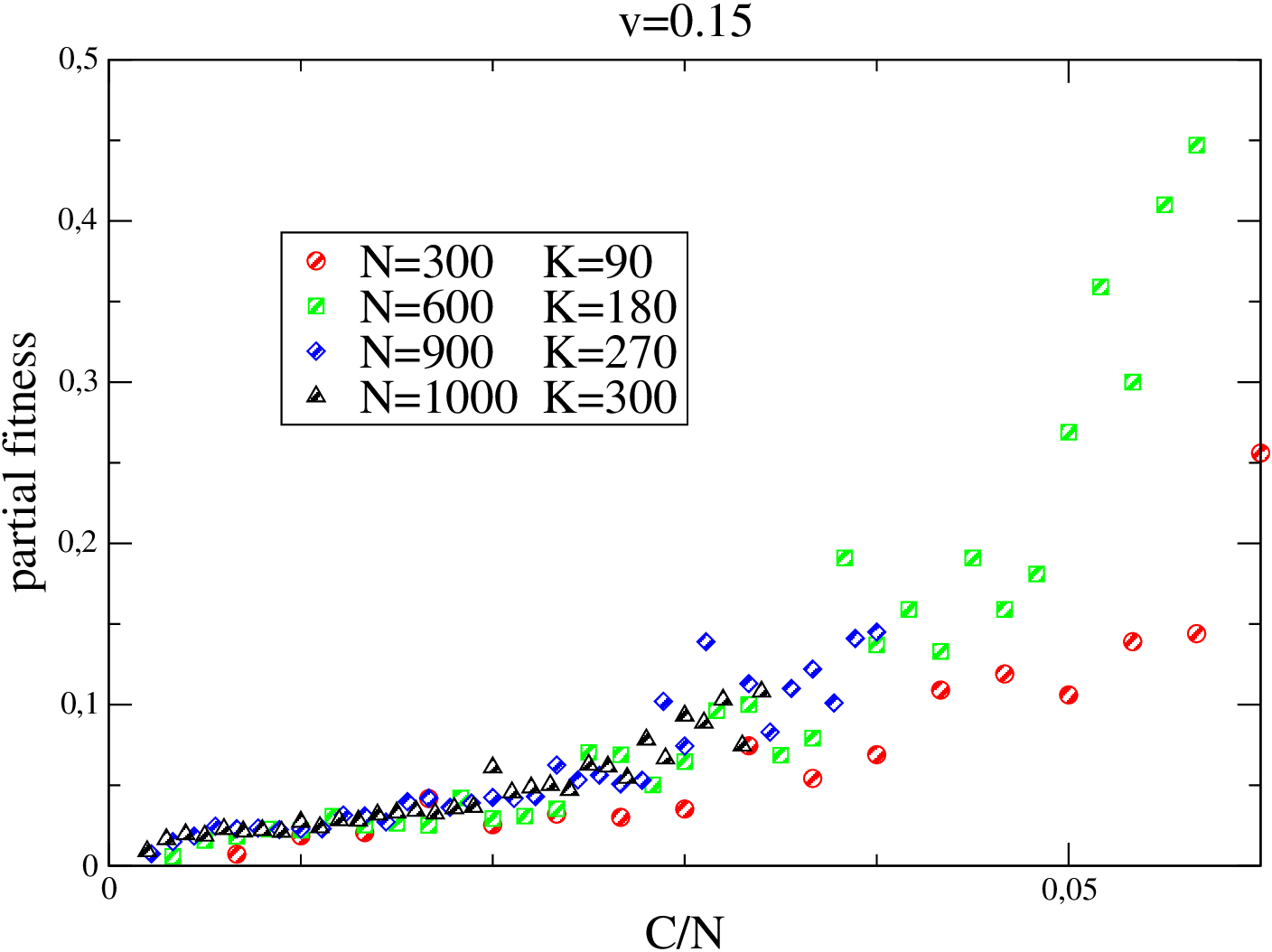}
\caption{\label{phasediag} Top left: phase diagram of the Hopfield model for $h^{\mathrm{ext}}=0$.
Points represents results of optimal solutions obtained by \textsc{cmaes} for various 
size $N$. Top right: Order parameters as a function of $\beta$  
if correlated states are correctly  detected by the guiding field.
Global (bottom left) and partial (bottom right) fitness
values of these solutions.}
\end{figure}

In that case, the terms corresponding to the local field $h_i$ becomes
irrelevant: their contribution to the energy per variable is then
$O(1/N)$. Hence the mean field limit is directly described by the
Hopfield model at inverse temperature
\[
\beta \egaldef \frac{4\alpha v}{\eta}.
\]
Let us simply describe the phase diagram $(T,\eta)$ (see
Figure~\ref{phasediag}) obtained in~\cite{AmGuSo3} for binary
$\xi_i\in\{-1,1\}$. When $C$ is macroscopic, the mixture acts in part
as a decorrelated random noise on the $J_{ij}$, so that a spin glass
phase, characterized by the Edwards-Anderson order parameter
\[
q \egaldef \frac{1}{N}\sum_{i=1}^N {\mathbb E}_\xi\Bigl({\mathbb E}_s
\bigl(s_i\vert\{\vec\xi_i\}\bigr)^2\Bigr),
\]
may develop and compete with the pure states encountered at finite
$C$. Except for a finite number of components $c=1,\ldots,s$, with
which a finite overlap may persist in the thermodynamic limit, 
the order parameter $\mu_c$ is otherwise of order $O(1/\sqrt N)$ for $c>s$ and
\[
r = \eta^{-1}\sum_{c>s}{\mathbb E}_{\xi}\left[\Bigl(\frac{1}{N}\sum_{i=1}^N
{\mathbb E}_s\bigl(s_i\xi_i^c\vert\{\vec\xi_i\}\bigr)\Bigr)
\Bigl(\frac{1}{N}\sum_{i=1}^N {\mathbb E}_s\bigl(s_i\xi_i^c\vert\{\vec\xi_i\}\bigr)\Bigr)\right],
\]
which represents the mean square of the global overlap with these
components, also introduced in~\cite{AmGuSo3}  may acquire a finite value. In presence of an external
field $h_i^{\mathrm{ext}} = \vec h\centerdot \vec \xi_i$ correlated with
the patterns, the mean-field equations of Amit, Gutfreund and Sompolinsky read
\begin{align}
\vec \mu &= {\mathbb E}_{\xi,z} \left[\vec \xi
\tanh\Bigl(\beta\bigl(\sqrt{\eta r}z+\vec\xi\centerdot(\vec\mu+\vec h)\bigr)\Bigr)\right],
\label{eq:mf1}\\[0.2cm]
q &={\mathbb E}_{\xi,z} \left[
\tanh^2\Bigl(\beta\bigl(\sqrt{\eta r}z+\vec\xi\centerdot(\vec\mu+\vec h)\bigr)\Bigr)\right],
\label{eq:mf2}\\[0.2cm]
r &= q/(1-\beta+\beta q)^2\label{eq:mf3},
\end{align}
where $z\sim\mathcal{N}(0,1)$ and where $\vec \mu$, $\vec \xi$ and
$\vec h$ are $s$-components vectors, if one assume the ground state to
be a state correlated to $s$ components of the mixture. For $\vec h =
0$, the phase diagram contains three phases, depending on the value of
$T=1/\beta$:
\begin{itemize}
\item the paramagnetic phase for $T> T_g$,
\item the spin glass phase for $T_c <T< T_g$,
\item the ferromagnetic phase for $T < T_c$, with spin configurations correlated with one
of the mixture component (Mattis states).
\end{itemize}
These are separated by two phase transition lines $T_g(\eta)$ (second
order) and $T_c(\eta)$ (first order).  An additional line $T_M(\eta)$
corresponds to the apparition of the Mattis states as metastable
states for $T_c<T<T_M$ before they become ground states for $T<T_c$.

Coming back to our inverse problem of finding the most accurate model
for inferring the underlying mixture distribution, the parameter
$\alpha$ allows us to tune $\beta$ to the most adequate value. For
this simplified formulation ($\xi_i^c\in\{-1,1\}$),  from the
definition (\ref{def:op}) of the order parameter and the definition (\ref{def:xi})
of $\xi_i^c$, we see that the
requirement is basically to tune $\beta$ such that the global optimum
corresponds to Mattis states with overlap
\begin{equation}\label{eq:request}
\mu = 2\sqrt v.
\end{equation}

\subsection{Mean-field decimation curves}
When the decimation procedure, described in Section~\ref{experiment}, 
is performed, the various indicators $R(\rho)$, $E(\rho)$ or $D_{\mathrm KL}(\rho)$
taken as functions of  $\rho$ give us a set of decimation curves, which 
we want to analyse in the mean-field regime.
When some variables are observed, the mean-field equations describing
the statistical behaviour of the hidden variables are simply obtained
by adding to their local field the field exerted by the observed
variables. Let $\rho$ be the fraction of observed variables, and
$\{s_i^*,i=1,\dots,\rho N\}$ the corresponding set. These variables
are correlated to one of the underlying component mixture, which we
choose to be $c=1$ by convention.  The reduced system consists then of
the $M = (1-\rho)N$ hidden variables, $\{s_i,i=1\ldots M\}$. To
simplify the discussion, we also assume that the connectivity in this
set is reduced in the same proportion to $(1-\rho)K$, which is
effectively the case on a complete graph. The external local field
experienced by any hidden variable $i$ now reads
\begin{align*}
h^{\mathrm{ext}}_i(\rho) &\egaldef h_i + \sum_{j^*\in i} J_{ij}s_j^*\\[0.2cm]
&=\frac{2\sqrt{v}}{\beta}\xi_i + \frac{\eta}{2}\bigl(\sum_{j\in i}\xi_{ij}
s_j^*- 2\sqrt v\sum_{j\in i}\xi_j\xi_{ij}\bigr) + 
K O\bigl(\frac{1}{C^{3/2}}\bigr),
\end{align*}
with $J_{ij}$ and $h_i$ given (\ref{eq:limcoup1}) and
(\ref{eq:limcoup2}). In the thermodynamic limit with $C=\eta K$, a
relevant term survives in $h_i(\rho)$ because of the correlations of
the $s_i^*$ with one of the mixture components (the first one by convention),
\[
{\mathbb E}_s\Bigl(\sum_{j\in i}\xi_{ij}s_j^*\Bigm\vert \xi\Bigr) = \frac{2\rho\sqrt v}{\eta}(\xi_i^1-\xi).
\]
As a result, keeping only the relevant term yields
\[
h_i^{\mathrm{ext}}(\rho) = 2\rho\sqrt v\xi_i^1 + O\bigl(\frac{1}{\sqrt C}\bigr).
\]
\textbf{For \boldmath$\rho=1$}: the single variable marginals (called the beliefs) are directly obtained from  
$h^{\mathrm{ext}}$ in this limit. To evaluate the prediction error, we have then simply to compare 
\[
\hat p_i^1(s_i=s) = \frac{1}{2} + s\sqrt v\xi_i^1. 
\]
with the corresponding limit belief,
\begin{equation}\label{def:belief}
p_i(s_i=s) = \frac{1}{2}\bigr(1 + s\tanh (\beta h_i^{\mathrm{ext}}(1))\bigl).
\end{equation}
After some algebra, we find (for $C\gg1$ and when the $\xi\in\{-1,1\}$
are binary) the following expression of the $D_{\mathrm{KL}}$ error,
\begin{equation}\label{eq:DKL}
D_{\mathrm{KL}}(p_i,\hat p_i^1) = 
\bigl(\frac{1}{2}+\sqrt w\bigr)\log\frac{1+2\sqrt w}{1+2\sqrt v}
+\bigl(\frac{1}{2}-\sqrt w\bigr)\log\frac{1-2\sqrt w}{1-2\sqrt v}
+O\bigl(\frac{1}{\sqrt C}\bigr),
\end{equation}
with $2\sqrt w = \tanh(\beta \sqrt v)$, so that the error vanishes when 
\[
2\sqrt v = \tanh(\beta\sqrt v).
\]
\textbf{For intermediate values of \boldmath$\rho$\,}: the mean field equations are still
valid after replacing $\beta$ by $(1-\rho)\beta$, $\eta$ by
$\eta/(1-\rho)$. The belief may be parametrized as in
(\ref{def:belief}) by a local field, which statistical ensemble is now
represented by the following stochastic variable
\begin{align*}
h(\rho) &= h^{\mathrm{ext}}(\rho) +\xi(1-\rho)\mu + \sqrt{(1-\rho)r\eta}z,\\[0.2cm]
&= \xi\bigr((1-\rho)\mu+2\rho\sqrt v\bigl) + \sqrt{(1-\rho)r\eta}z,
\end{align*}
where $\xi$ has variance 1, $z \sim \mathcal{N}(0,1)$, and $r$ is such
that ${\mathbb E}_{\xi,z}[\tanh^2(\beta h_i)] = q$. The mean
Kullback-Leibler distance with the reference belief $\hat p$ then
reads,
\begin{align}\label{eq:dkl}
D_{\mathrm{KL}}(p,\hat p) &= {\mathbb E}_{\xi,z}\Bigl(\beta(h-\hat h)\tanh(\beta h) 
+\log\frac{\cosh\beta\hat h}{\cosh\beta h}\Bigr),\nonumber\\[0.2cm]
&=  \beta\mu[(1-\rho)\mu + 2\rho\sqrt v] +\beta^2r\eta(1-\rho)(1-q)\nonumber\\[0.2cm]
&+{\mathbb E}_{\xi,z}\Bigl[\log\frac{1-\tanh^2 (\beta h)}{1-4v\xi^2}
-\atanh(2\xi v)\tanh(\beta h)\Bigr].
\end{align}
For binary variables $\xi\in\{-1,1\}$, we recover (\ref{eq:DKL}) when
$\rho=1$ with $\mu=2\sqrt w$. In this special case it is in fact
tempting to tune $\alpha$ such that the requirement (\ref{eq:request})
is fulfilled for any $\rho$. Tuning the function $\alpha(\rho)$
amounts to find $\beta$ such that
\begin{align*}
2\sqrt v  &= {\mathbb E}_z\left[
\tanh\Bigl(\beta\bigl(\sqrt{(1-\rho)\eta r}z
+2\sqrt v\bigr)\Bigr)\right]\\[0.2cm]
q  &= {\mathbb E}_z\left[
\tanh^2\Bigl(\beta\bigl(\sqrt{(1-\rho)\eta r}z
+2\sqrt v\bigr)\Bigr)\right],\\[0.2cm]
r &= \frac{q}{\bigl(1-\beta(1-\rho)(1-q)\bigr)^2},
\end{align*}
altogether with equation (\ref{eq:mf3}), when $\sqrt v$ and $\eta$ are fixed
parameters. 
Instead, when $\xi$ is continuously distributed, the resulting $D_{\mathrm{KL}}$ error 
is then a superposition of elementary distances, and has a strictly positive lower
bound.
\begin{figure}[ht]
\centering
\includegraphics*[width=0.49\linewidth]{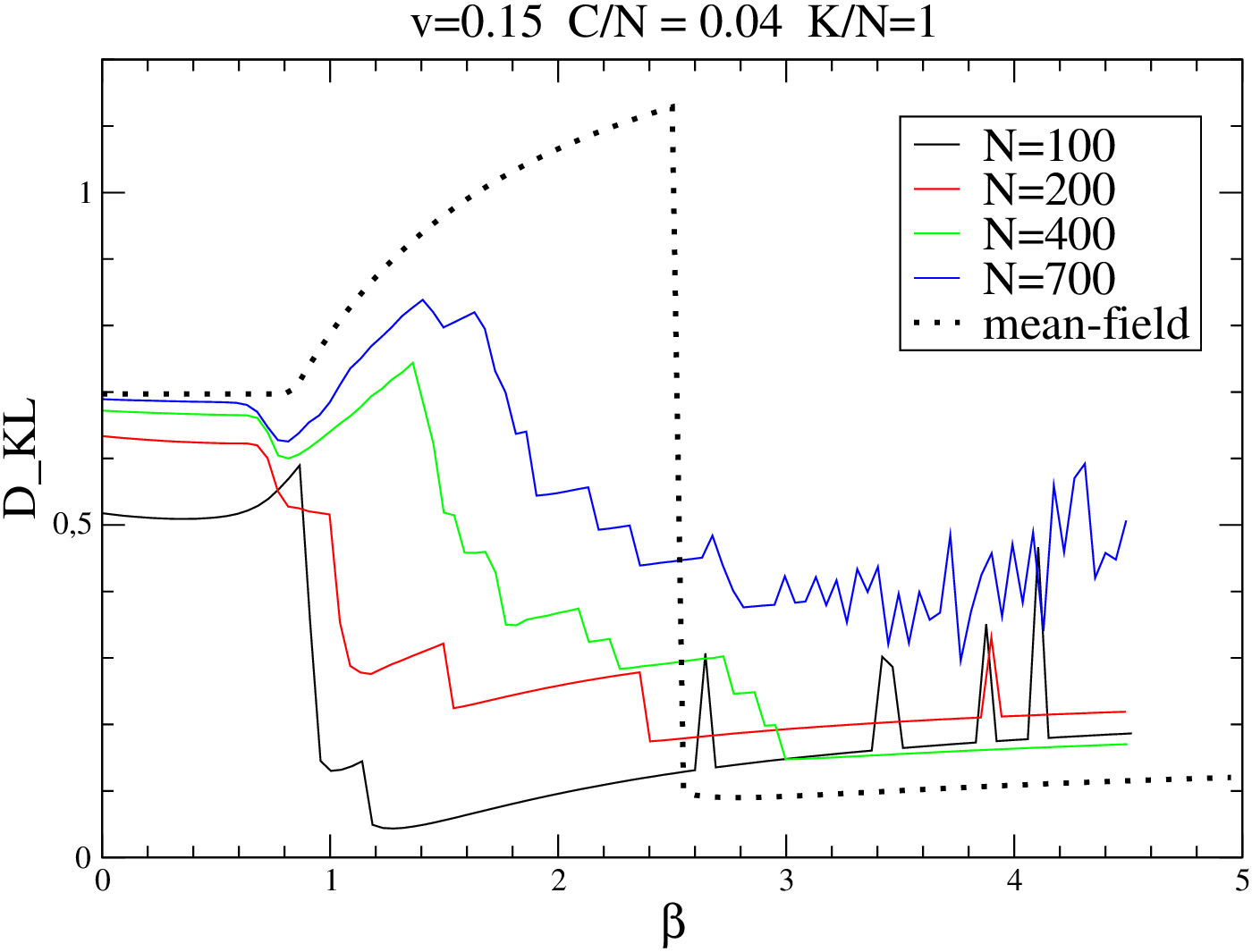}
\includegraphics*[width=0.49\linewidth]{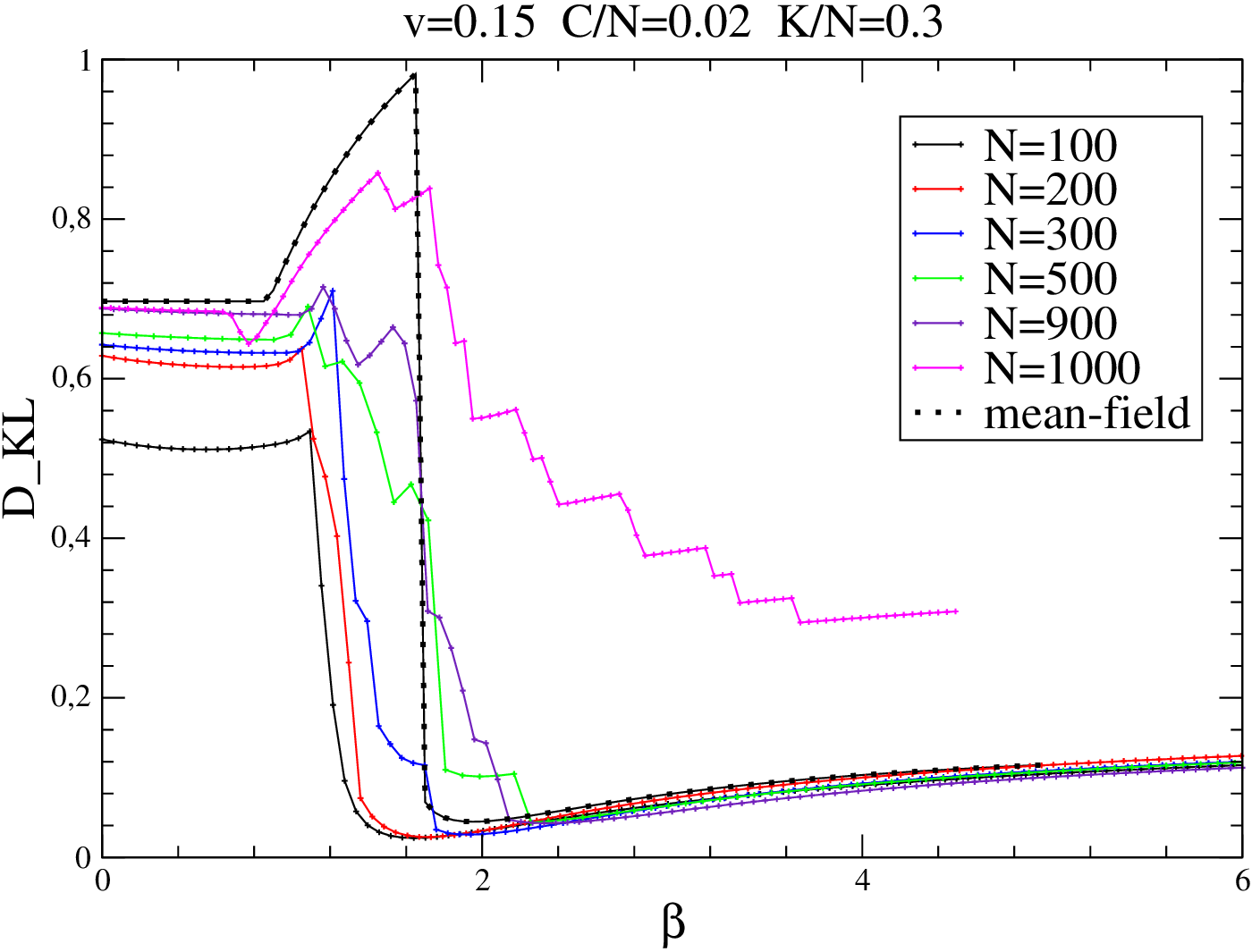}
\includegraphics*[width=0.49\linewidth]{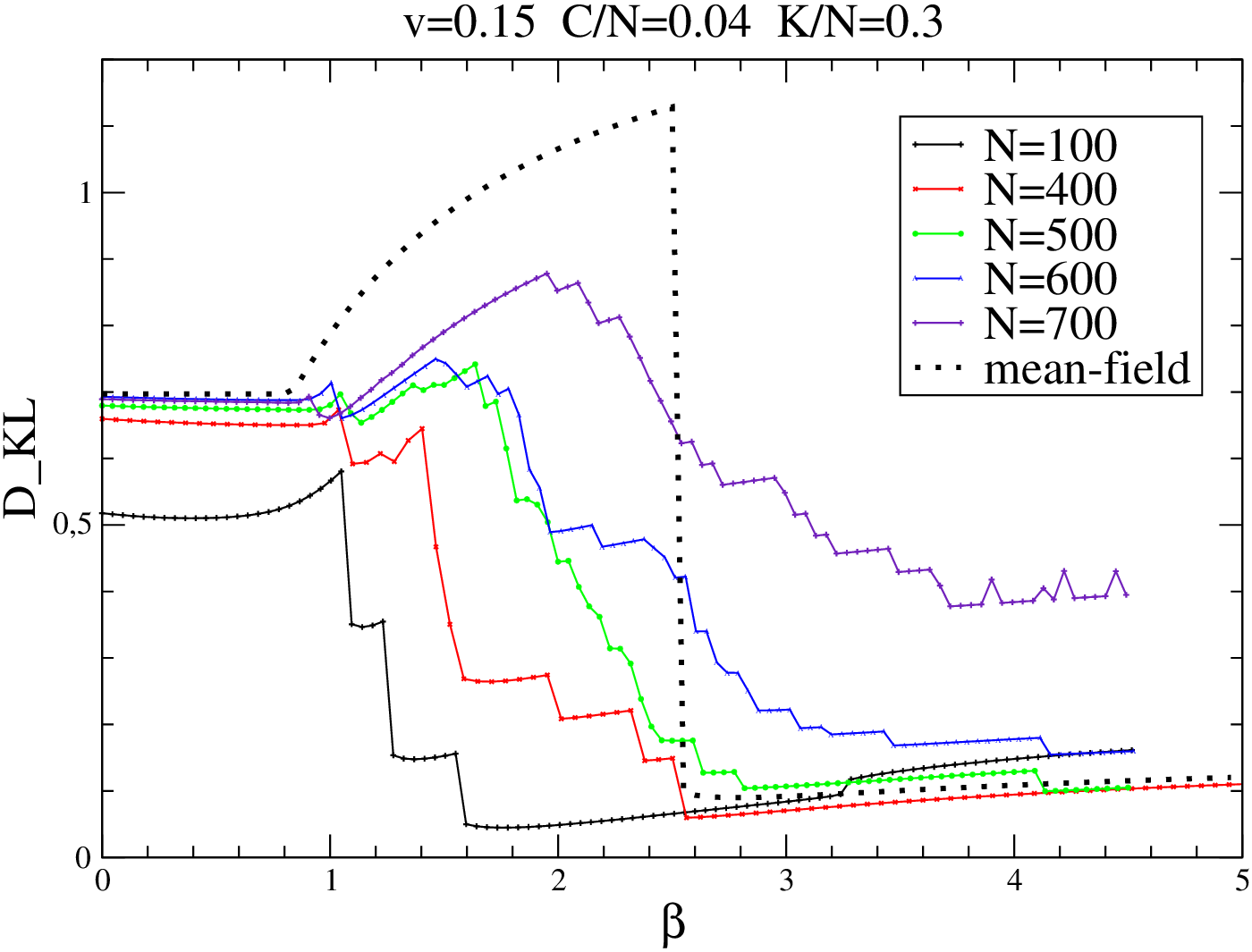}
\includegraphics*[width=0.49\linewidth]{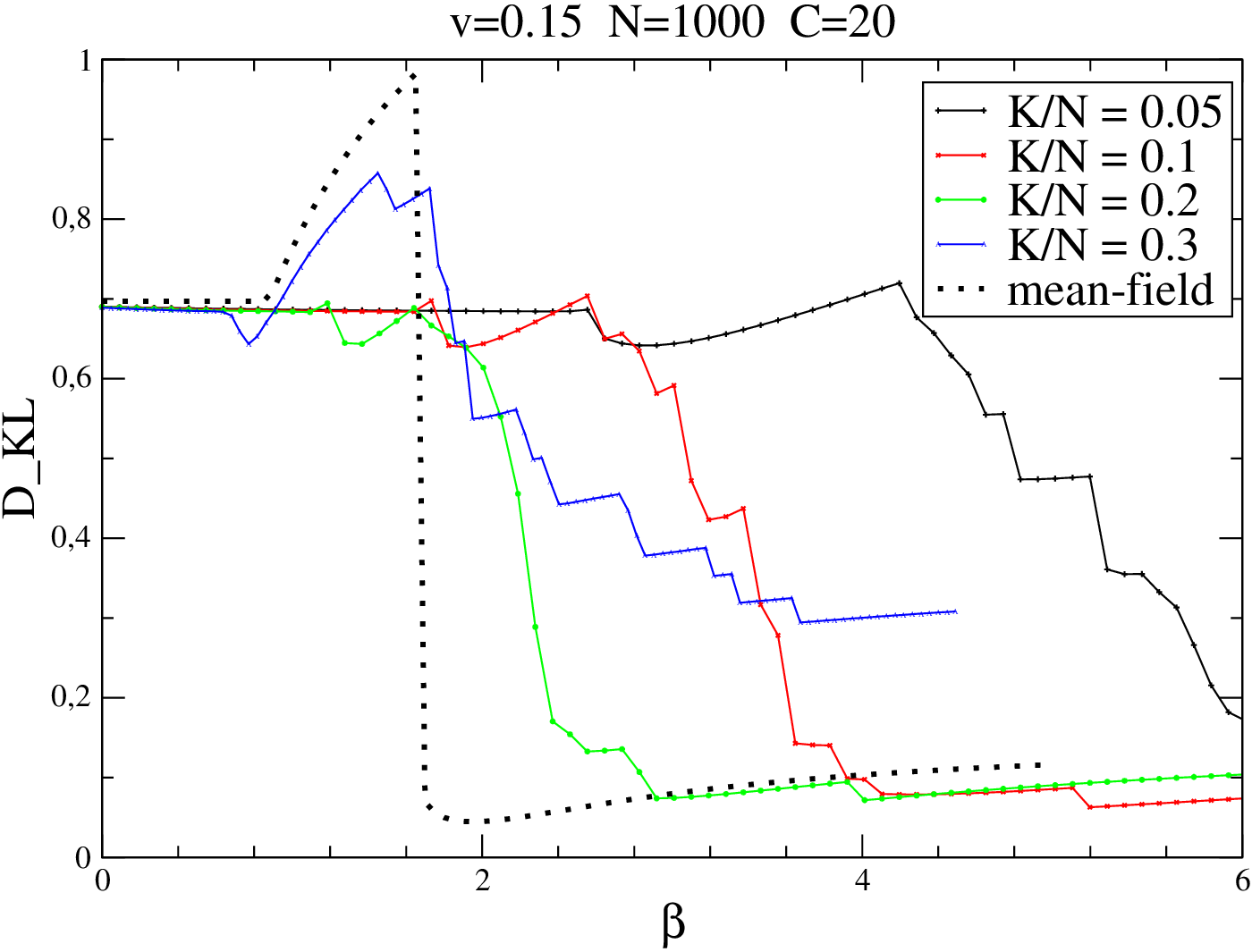}
\caption{\label{dkldebta} The Kullback-Leibler error as a function of $\beta$ 
obtained experimentally with an evanescent guiding field and their corresponding 
mean-field expectation (\ref{eq:dkl}). 
The top left plot shows the limitation due to the spin-glass phase.
Effects of the pruning procedure is shown on the other  plots.}
\end{figure}%
\begin{figure}[ht]
\centering
\includegraphics*[width=0.49\linewidth]{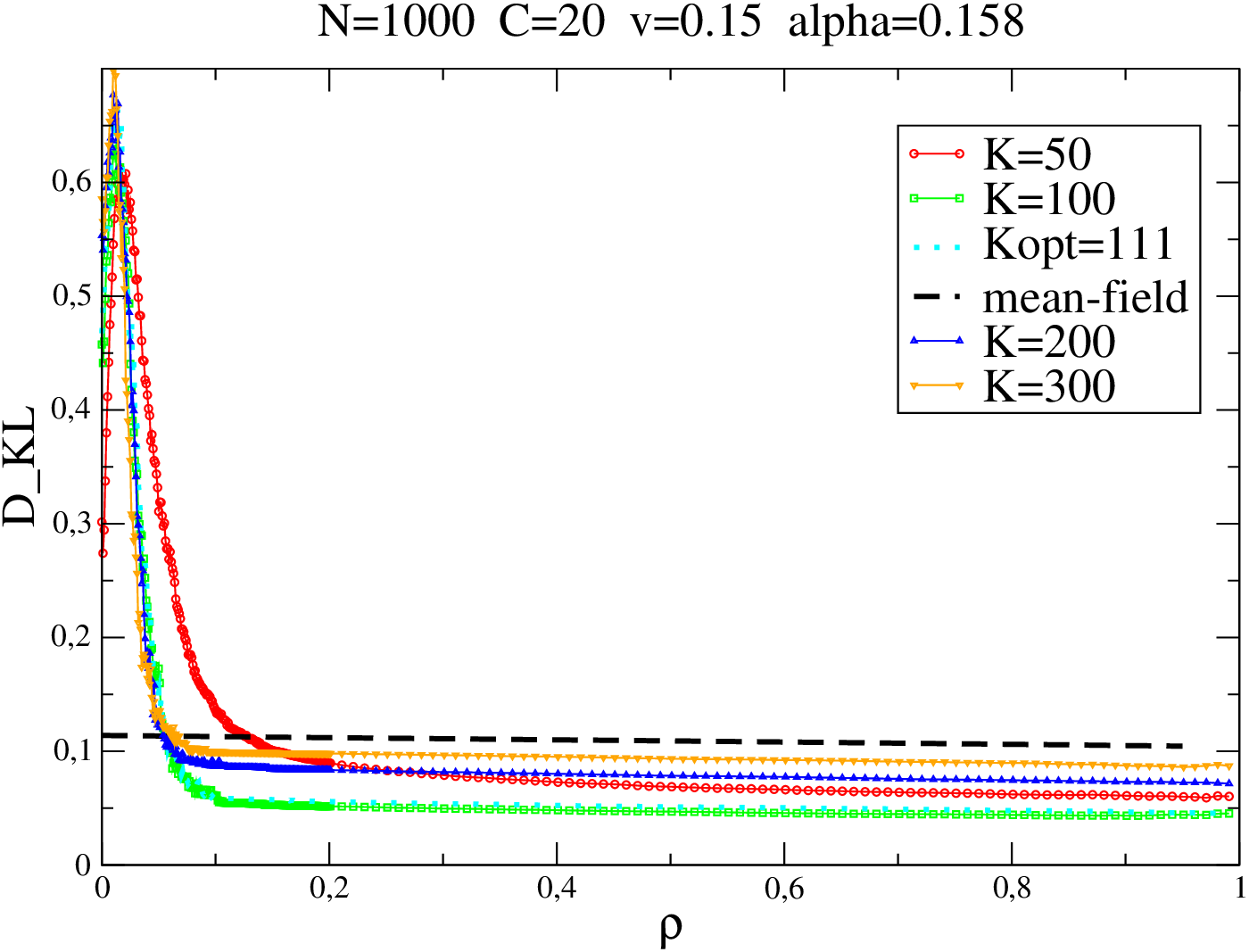}
\includegraphics*[width=0.49\linewidth]{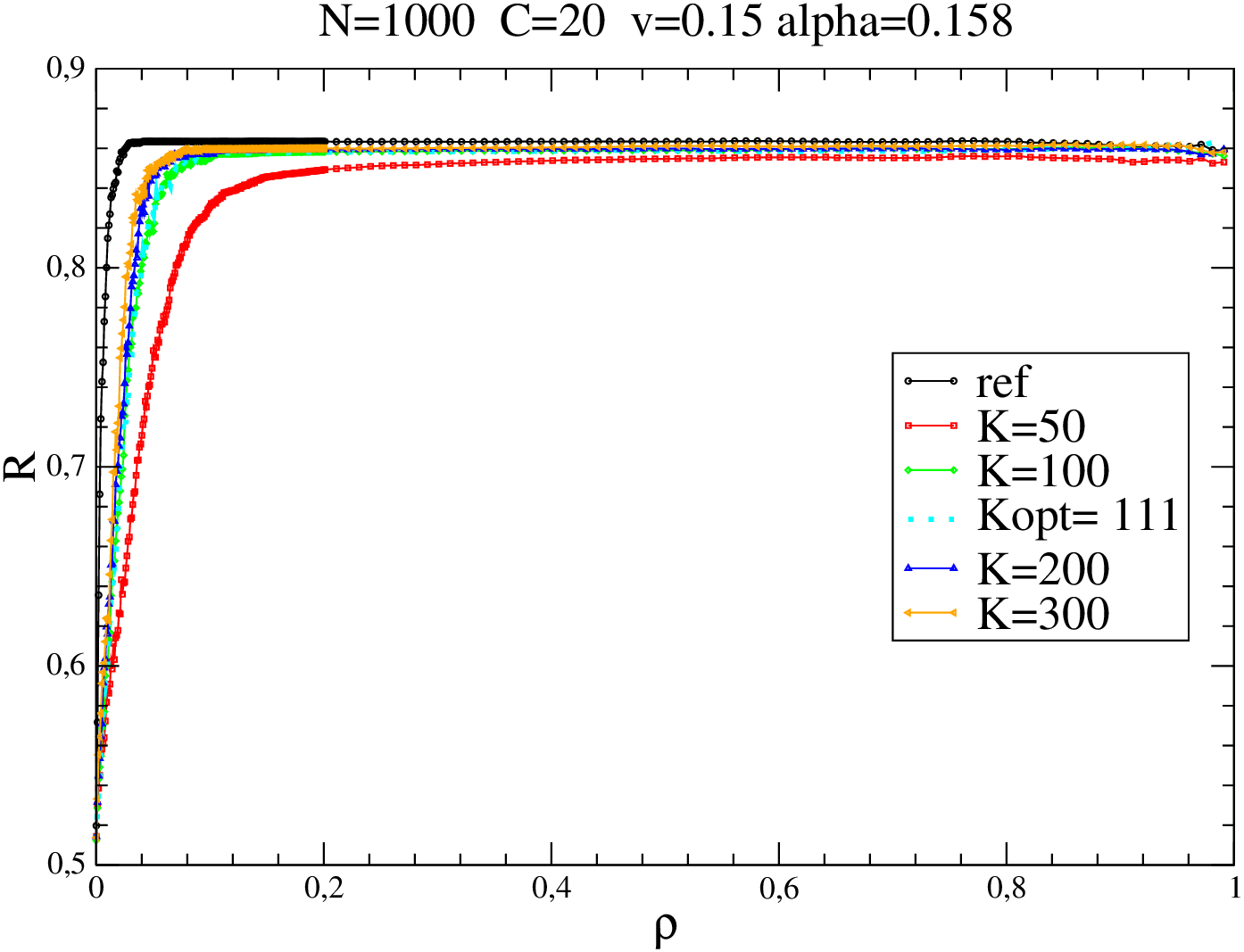}
\includegraphics*[width=0.5\columnwidth]{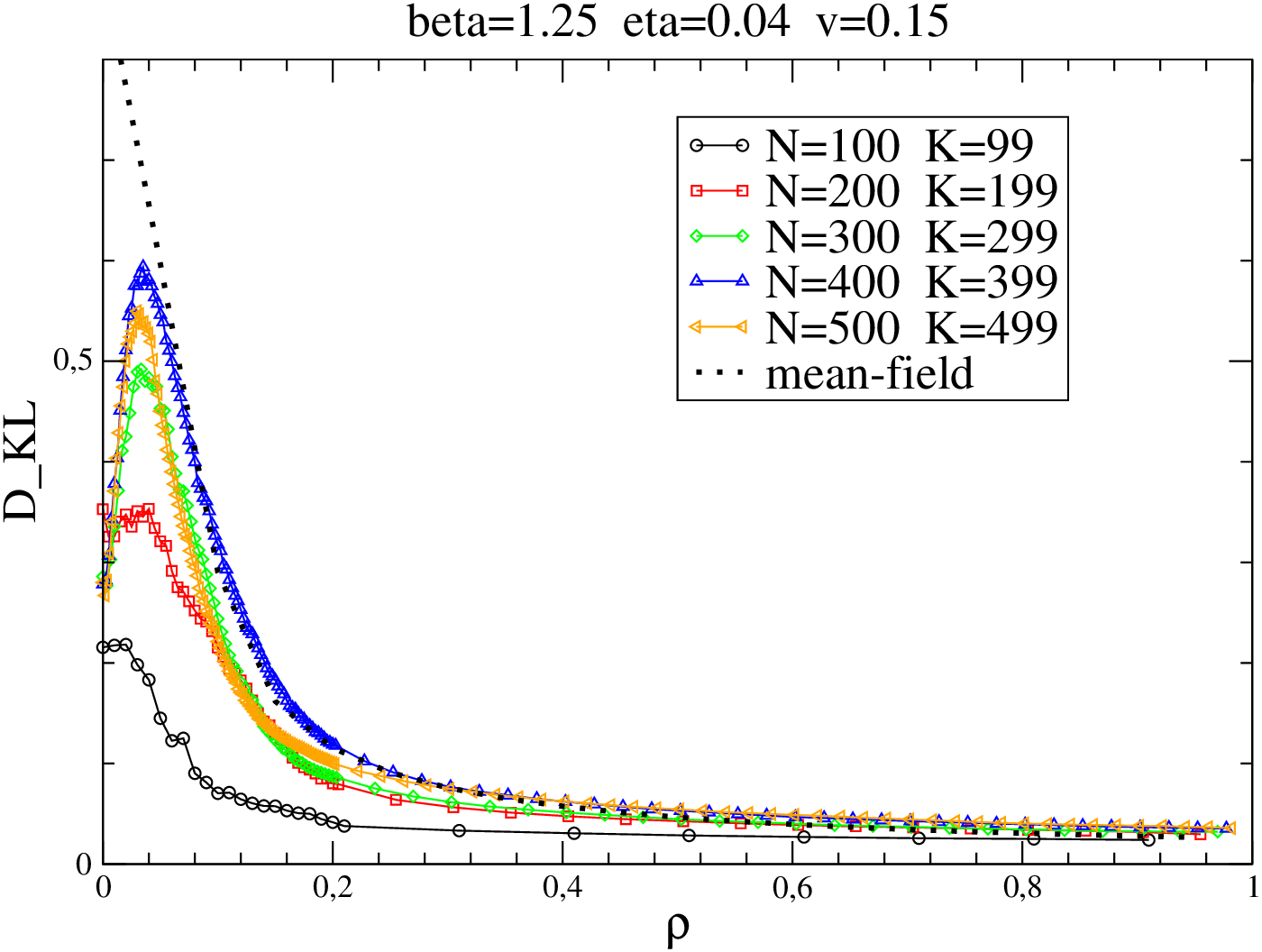}
\caption{\label{decim} 
Bottom: Experimental decimation curves of $D_{\mathrm{KL}}$ at fixed $\beta=1.25$ and $C/N=0.04$
for complete graphs, compared with their expected mean-field limit. Top: Effect
of pruning on the $D_{\mathrm{KL}}$ error (left) and on the prediction error (right)
for $C/N=0.02$ and $\beta=4.74$.}
\end{figure}%
 
\subsection{Comparison with experimental results}
The numerical results presented in Figures~\ref{phasediag}--\ref{optim100}   
are obtained by running \textsc{lpb}
on the experimental setting explained in Section~\ref{experiment}, 
performed with a fixed intermediate value of $v=0.15$, 
along with the inference model presented in Section~\ref{heuristic} and~\ref{strengthpar}. 

Consider first what is expected to 
happen, for small enough value of $C/N$, when correlated states are
searched with the help of a guiding field (see
Section~\ref{experiment}), while $T$ is decreased along a vertical
line on the phase diagram (see top left of Figure~\ref{phasediag}):
the spin-glass transition line $T_g$ is first encountered,
materialized by a sudden increase of $r$ and $q$ as well as $D_{\mathrm{KL}}$
(see top right Figure~\ref{phasediag}). The small amount of
information contained in the paramagnetic phase get simply screened by
the proliferation of spurious states, none of them being correlated
with the Mattis states ($\mu=0$). Then the line $T_M$ is passed
through, correlated states appears, which are expected to be detected
by the guiding field, so that $\mu$ acquire a non-zero value, while
$r$ decreases.
In practice, as seen from the top left Figure~\ref{dkldebta}, the spin
glass phase renders the guiding field ineffective when $N$ increases.
The pruning procedure cure partially this problem, but a trade-off has
to be found, as can be see from the bottom right
Figure~\ref{dkldebta}: the density of spurious states decreases when
the pruning increases, but phase transition lines get shifted in a way
that allows only highly polarized states to be present; as a result,
the lower bound of $D_{\mathrm{KL}}$ increases. Intermediate pruning threshold
have been actually found by the optimization procedure (see next
section) and the phase diagram remains approximately valid, as seen by
looking at the top right and bottom left of
Figure~\ref{dkldebta}\footnote{The true phase diagram after pruning is
  actually unknown to us, because the links are not chosen randomly.
  $N$ seems to be more appropriate than $K$ to define the temperature
  for intermediate values of the pruning (e.g.\ $K/N=0.3$)}. We
observe that the solutions remain close to the $T_M$ line in
Figure~\ref{phasediag}.
Concerning the decimation plots (Figure~\ref{decim}), comparison with
the mean-field limit differs at low density $\rho$ because of finite
size effects (top) and because of the spin-glass phase (bottom), which
prevents the \textsc{lbp} to converge faithfully to the ground states.
The saturation phenomena of the decimation curves, which occurs when
$\rho$ tends to 1, is reproduced correctly by the mean-field analysis.
One would expect the $D_{\mathrm{KL}}$ error to vanish as the number of
observed variables increases, but, as indicated by (\ref{eq:dkl}), we
have a superposition of $D_{\mathrm{KL}}$ errors, due to the dispersion in the
polarization of variables, which by definition cannot be made
arbitrarily small.  Still, Figure~\ref{decim} is an instance where an
efficient prediction is obtained with less than five percent of
observed variables, which could be is useful for real applications.

\section{Continuous parameter optimization}\label{optimization}
\begin{figure}
\centering
\includegraphics*[width=0.48\linewidth]{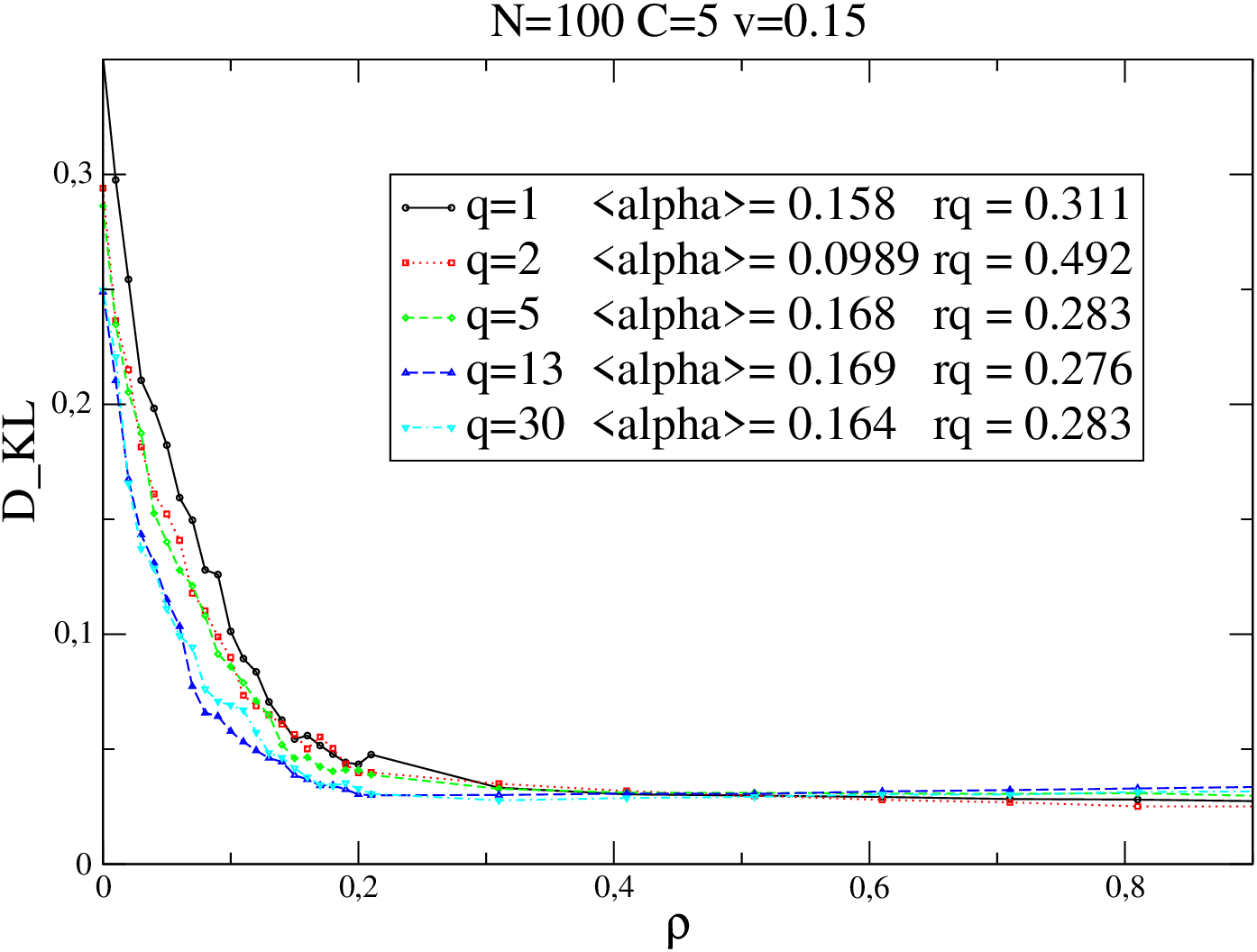}
\includegraphics*[width=0.5\linewidth]{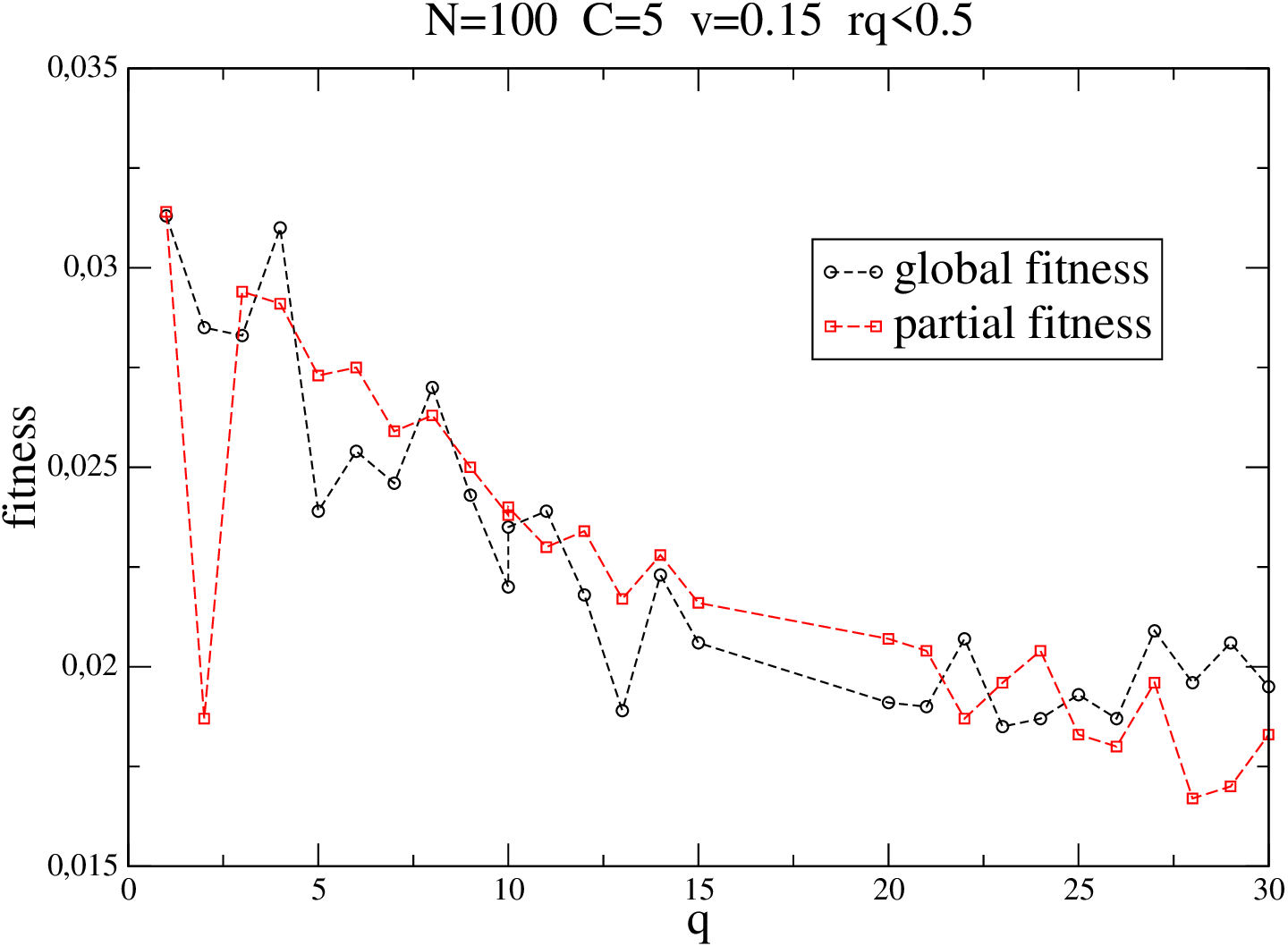}
\caption{\label{optim100} Optimization results for a problem
  with $N=100$ variables and $C=5$ components, when $q$
  is increased. The optimization is performed
  with the upper quantile $r_q$
  bounded to 0.5. The average single variable $D_{\mathrm{KL}}$ error as a function of
  $\rho$, the fraction of observed variables (left). Correlation
  between the global and partial fitness (right).    
 }
\end{figure}

The definition (\ref{def:lbpmeasure}) sets up a single parameter model
which, combined with the pruning procedure, is in fact a two parameter
model $\omega =(\alpha,r)$ where $r\in[0,1]$ is the fractions of edges
which are conserved.  The model could be straightforwardly extended by
associating a coefficient $\alpha_a$ to each factor node $a$. The
determination of the set $\{\alpha_a,a\in \F\}$ for optimizing the
model, would lead to a difficult continuous and combinatorial
optimization problem. Instead, assuming we have at hand a meaningful
criteria to sort the factor nodes, we may divide the distribution in a
certain number of parts $q$, delimited by a an increasing set of quantiles $\{r_i,i=0,\ldots,q\}$, 
with $r_0 = 0$ and $r_q \le 1$, each part associated to a parameter $\alpha_i$. As a result, given the
number of parts $q$, we have a $2q$ parameter model,
$\omega^{(q)}=(\alpha_1,\ldots,\alpha_q,r_1,\ldots,r_q)$, which is
well suited to continuous optimization, if $q$ is not too large
(typically less than $100$). This requires the definition of a
fitness function. We have conducted this program on the pairwise
model. The natural fitness function for this problem is obtained from
the decimation procedure explained in Section~\ref{experiment},
\[
F(\omega^{(q)}) \propto \int_0^1 d\rho(1-\rho)D_{\mathrm{KL}}(\rho),
\]  
where $\rho$ is the fraction of observed variables. This fitness
function is however quite costly, so we use a surrogate fitness function based
on the identifications of the fixed points:
\[
\tilde F(\omega^{(q)}_q) \propto \sum_{c=1}^C D_{\mathrm{KL}}^{(c)}(0).
\]  
where $D_{\mathrm{KL}}^{(c)}(0)$ represents the Kullback-Leibler
marginal distance of a driven fixed point (with help of the evanescent
guiding field introduced in Section~\ref{experiment}) to the
corresponding mixture component $c$ when all variable are hidden. This
surrogate fitness appears to be much less noisy and costly than the
original one, but still well correlated to it as can be seen in
Figure~\ref{optim100} (right). One can get an idea of the ruggedness
of the fitness landscape by simply looking at Figure~\ref{dkldebta}.
As a consequence we used  a stochastic optimization algorithm,
usually well suited choice for rugged fitness landscapes. The
optimizer chosen is the Covariance-Matrix-Adaptation
Evolution-Strategies (CMA-ES) \cite{Hansen2001}, where a population of
candidate solutions are sampled according to a multivariate normal
distribution, whose parameters (mean value and covariance matrix) are
adapted according to the feedback gathered along the optimization
procedure. The underlying idea for the adaptation mechanism is to
increase the probability of sampling better solutions. In the end of
the search procedure, the sampling distribution gives an estimate of
the local curvature of the objective function.

We have compared different
ways of sorting the edges based on the set of coupling $J_{ij}$ (see
preceding section), which somehow figure the amount of information
transmitted from one variable node to another one. Based on the
electric network analogy (see e.g.\ \cite{Grimmet}), we consider the
following different sorting criteria:
\begin{itemize}
\item simple sorting,  
\item absolute conductance sorting,  
\item relative conductance sorting.
\end{itemize}
We expect these to capture different properties of the underlying
factor graph. The simple sort is based on the value of $\vert
J_{ij}\vert$ for each edge $(i,j)\in\E$. The absolute conductance sort
amounts to reweight these couplings $J_{ij}$ by the fraction of
weighted spanning tree (WST) containing the edge $(i,j)$, while the
relative conductance sorting take into account this fraction solely
(the spanning trees are weighted with these $\vert J_{ij}\vert$).
Deceptively, the simple sorting procedure yields the better results.
So if there exists a smarter way of sorting the links, we might find it
hopefully by analyzing the mean field equation on a pruned graph,
which are not established yet. Anyway, the example shown on
Figure~\ref{optim100} indicates that the optimization works when using
this simple sorting procedure. In this example, the global error is
decreased by $40\%$ with a $13$ quantiles parameters model with
respect to the single parameter model (Figure~\ref{optim100}, right).
In addition, the improvements occur in the region of interest, that is
when $\rho<0.2$ (Figure~\ref{optim100}, left).

\section{Comparison with other approaches and perspectives}\label{comparison}

The model we propose shares some common points with the
tree-reweighted belief propagation algorithm described in~\cite{Wai06}
and with the fractional belief propagation scheme~\cite{WiHe}.  The
Bethe approximation (\ref{def:bfe}) is a particular case of a general
set of variational region based free energy
approximations~\cite{YeFrWe3}. Introducing for each variable and
factor node the energies and entropies,
\begin{align*}
E_i \egaldef -\sum_{x_i}\bel_i(x_i)\log \phi_i(x_i)&\qquad E_a \egaldef
-\sum_{\x_a}\bel_a(\x_a)\log \psi_a(\x_a),\\[0.2cm]
H_i \egaldef -\sum_{x_i}\bel_i(x_i)\log \bel_i(x_i)&\qquad H_a \egaldef
-\sum_{\x_a}\bel_a(\x_a)\log \bel_a(\x_a),
\end{align*}
and considering only the region associated to the factors, 
a general approximation is obtained by introducing different counting 
numbers for the average energy and entropy,
\begin{align}\label{def:gbfe}
F(\bel) = \sum_a (e_a E_a - h_a H_a) + \sum_i (e_i E_i - h_i H_i)
\end{align}
The coefficients corresponding to the fractional belief propagation
approach of~\cite{WiHe} are 
\[
e_a = 1\qquad e_i= 1\qquad h_i = 1-\sum_{a\ni i}h_a,
\]
where the $h_a$ are arbitrary real coefficients.

Concerning the tree reweighted free energy of~\cite{Wai06}, which is
defined for a pairwise factor graph, as noted in~\cite{WeYaMe} the coefficients read
\[
e_{ij} = 1\qquad e_i= 1\qquad h_i = 1- \sum_{j}h_{ij},
\]
where $h_{ij}\in[0,1]$ represents the probability that edge $(i,j)$
appears in a spanning tree of $\G$, chosen randomly under some given
measure on the set of spanning trees. It is too a sub-case of
fractional belief propagation.

Our choice instead amounts to consider the parametrization
\[
e_i= 1   \qquad h_i = 1-d_i\qquad h_a=1,
\]
while $e_a$ are arbitrary positive coefficients, noted $\alpha_{ij}$,
with the convention (\ref{def:lbpmeasure}) for $\phi$ and $\psi$. 

It is however not this slight modification of the search space of
approximate variational free energy that characterizes our approach,
but rather the variational framework. In our case, we purposefully
choose a non convex framework, because we want to allow many
belief-propagation fixed points to be present. Conversely,~\cite{WiHe}
and~\cite{Wai06} strive at finding a convex variational free energy
approximation.  Further work is needed, possibly by extending the
search to the full variational space corresponding to the set of
coefficients $(e_a,e_i,h_a,h_i)$, to see which type of parametrization
is best adapted to our problem. Let us simply note for the moment that
counting coefficients $h_a\ne 1$ and $h_i\ne 1-d_i$ yield some feed-back
in the definition of the messages (see Appendix), which is precisely
what this message passing procedure is supposed to avoid for obtaining
fast convergence. Nevertheless, it would be interesting to see whether
the measure on weighted spanning trees deduced from the strength of
the coupling constants may be used to define a well suited tree
reweighted approximation.

The main observation of this work, namely that a mixture of well
separated probabilistic states may be efficiently encoded and decoded
in a multiple set of \textsc{lbp} fixed points, deserves further
developments, both from the practical and theoretical point of view.
The analysis of the mean field theory could be extended to understand
better how graph pruning affects the equations. More generally,
understanding better the influence of the graph structure on the mean
field equation could yield as a byproduct an optimal way of sorting
the edges for the optimization procedure. Further work is also needed
regarding the effect of the factor graph on the storage capacity, when
not restricting ourselves, as in the present study, to a pairwise
factor graph.  While trying to optimize the number of probabilistic
patterns that may be encoded, we have at the same time to restrain the
connectivity of the graph, so that the advantage of using a fast
message procedure is preserved: a proper trade off has to be found. In
addition, the connection with the Hopfield model helps us to assess
the limitation due to spin glass effects, and developments in the
field of neural networks should help us to limit this drawback.

\paragraph{Acknowledgments} This work was supported by the French National Research Agency (ANR) grant N° ANR-08-SYSC-017.

{\small \bibliography{refer}}
\bibliographystyle{unsrt}

\newpage
\appendix
\section{Appendix: Generalizations to belief propagation algorithm}
We adapt here the reasoning of \cite{YeFrWe3} to the free energy of 
Section~\ref{comparison}. The function that has to be studied to minimize the 
generalized Bethe free energy (\ref{def:gbfe}) reads
\begin{align}\label{def:freeEn}
\mathcal{F}_{\lambda\gamma}(\bel) &= - \sum_{a,\x_a} \bel_a(\x_a)\log\frac{\psi_a(\x_a)^{e_a}}{\bel_a(\x_a)^{h_a}} 
- \sum_{i,x_i} \bel_i(x_i)\log\frac{\phi_i(x_i)^{e_i}}{\bel_i(x_i)^{h_i}}\nonumber\\
&+ \sum_{\substack{i,a\ni i\\ x_i}} \lambda_{ai}(x_i)\bigl(\bel_i(x_i) - \sum_{\x_{a\setminus i}}\bel_a(\x_a) \bigr)
- \sum_i \gamma_i\bigl(\sum_{x_i} \bel_i(x_i) -1\bigr),
\end{align}
with $\{\lambda_{ai}\}$ a set of Lagrange multipliers attached to each
link, to insure compatibility conditions between joint beliefs and
single beliefs, and $\{\gamma_i\}$ a set destined to enforce single
beliefs normalization. The stationary points read
\[
\begin{cases}
\DD \bel_a(\x_a) &= \psi_a(\x_a)^{e_a/h_a} \exp\Bigl(\frac{1}{h_a}\sum_{i\in a}\lambda_{ai}(x_i) - 1\Bigr),\\[0.2cm] 
\DD \bel_i(x_i) &= \phi_i(x_i)^{e_i/h_i} \exp\Bigl(\frac{1}{h_i}\bigl(\gamma_i-\sum_{a\ni i}\lambda_{ai}(x_i) 
\bigr)-1\Bigr).
\end{cases}
\]

At this stationary point, the generalized Bethe free energy reads
\begin{align*}
\mathcal{F}(\bel) 
&=  - \sum_{a,\x_a} \bel_a(\x_a)\Bigl[h_a -  \sum_{i\in a}\lambda_{ai}(x_i)\Bigr]
- \sum_{i,x_i} \bel_i(x_i)\Bigl[h_i + \sum_{a\ni i}\lambda_{ai}(x_i)-\gamma_i\Bigr]\\
&=  \sum_i \gamma_i - \sum_a h_a - \sum_i h_i.
\end{align*}
and one can write
\[
\prod_a \psi_a(\x_a)^{e_a}\prod_i\phi_i(x_i)^{e_i} = 
\prod_a \bel_a(\x_a)^{h_a}\prod_i \bel_i(x_i)^{h_i}
e^{-\mathcal{F}(\bel)}, 
\]

The compatibility constraint between the single variable beliefs $\bel_i$ and
factor beliefs $\bel_a$ yields for $i\in a$
\begin{equation}\label{eq:constraint}
\sum_{x_{a\setminus i}} \psi_a(\x_a)^{e_a/h_a}\prod_{j\in a}
n_{j\to a}(x_j)^{1/h_a} \propto 
\frac{\phi_i(x_i)^{e_i/h_i}}{\prod_{a'\ni i} n_{i\to a'}(x_i)^{1/h_i}} 
\end{equation}
with the usual definition, although slightly different from~(\ref{urulesn}),
\begin{equation}
n_{i\to a}(x_i) \egaldef \exp(\lambda_{ai}(x_i)).\label{urulesn2}
\end{equation}

A simple way of getting a mapping suitable for an iterative algorithm
is to isolate the term $n_{i\to a}(x_i)$ to the left of the equation
\begin{align*}
n_{i\to a}(x_i)^{-(1/h_a+1/h_i)} 
 \propto& \sum_{x_{a\setminus i}} \Bigl[\psi_a(\x_a)^{e_a}
   \prod_{j\in a\setminus i} n_{j\to a}(x_j)\Bigr]^{1/h_a}\\
 &\times \Bigl[\phi_i(x_i)^{-e_i}\prod_{\substack{ a'\ni i\\ a'\ne a}} n_{i\to a'}(x_i)\Bigr]^{1/h_i}.
\end{align*}
This relation yields a new message passing algorithm that would be a
close cousin of the \textsc{lbp} algorithm; the properties of this new
algorithm have not been investigated yet. 

In order to obtain something that is closer to the original algorithm,
we define a new set $\{m\}$ of messages by the relation
\[
m_{a\to i}(x_i) \egaldef n_{i\to a}(x_i)^{-1/h_a}\prod_{a'\ni i}n_{i\to a'}(x_i)^{-1/h_i},
\]
and rewrite~(\ref{eq:constraint}) as
\begin{equation}\label{eq:gupdate1}
m_{a\to i}(x_i) 
 \propto \sum_{x_{a\setminus i}} \Bigl[\psi_a(\x_a)^{e_a}
   \prod_{j\in a\setminus i} n_{j\to a}(x_j)\Bigr]^{1/h_a}\times \phi_i(x_i)^{-e_i/h_i}.
\end{equation}

This relation will produce a \textsc{lbp}-like algorithm if we invert the definition of $\{m\}$. To this end, we write the identity
\[
\sum_{a'\ni i} h_{a'}\log\bigl(m_{a'\to i}(x_i)\bigr) = -\sum_{a'\ni i} \frac{h_i+\sum_{b\ni i} h_b}{h_i}
\log \bigl(n_{i\to a'}(x_i)\bigr),
\]
from which the following relation can be obtained
\begin{align}
 \log\bigl(n_{i\to a}(x_i)\bigr) 
&= -h_a\log\bigl(m_{a\to i}(x_i)\bigr)+\frac{h_a}{h_i+\sum_{b\ni i}h_b}\sum_{a'\ni i}h_{a'}\log\bigl(m_{a'\to i}(x_i)\bigr).\label{eq:gupdate2}
\end{align}

Equations (\ref{eq:gupdate1})--(\ref{eq:gupdate2}) yield the updates rules in this generalized setting. In the case of fractional belief propagation, (\ref{eq:gupdate2}) reduces to  
\[
\log\bigl(n_{i\to a}(x_i)\bigr) 
= -h_a\log\bigl(m_{a\to i}(x_i)\bigr)+
h_a\sum_{a'\ni i}h_{a'}\log\bigl(m_{a'\to i}(x_i)\bigr)
\]
The ordinary \textsc{lbp} scheme corresponds to $h_a=1$ and
$h_i=1-d_i$. Note that, contrary to the fractional belief propagation
algorithm and to the tree-reweighted algorithm, there is no feedback
term apparent in the r.h.s.\ of (\ref{eq:gupdate1}). This property
ensures the independence of the messages in absence of loops.

However, the definition in (\ref{eq:gupdate1}) contains a feedback at
second order since $m_{a\to i}$ depends of $m_{a\to j}$ for $j\ne i$,
which themselves have been computed from the former value of 
$m_{a\to i}$. This can be avoided only when
\[
 -h_a\Bigl(1-\frac{h_a}{h_i+\sum_{b\ni i}h_b}\Bigr)=0,
\]
that is, $h_a=h$ and $h_i=(1-d_i)h$ for some value of $h$. This
setting is equivalent to normal \textsc{lbp}.
 
\end{document}